%% file: main.tex
\documentclass[10pt,twocolumn,letterpaper]{article}
\usepackage[pagenumbers]{cvpr} %

\usepackage{graphicx}
\usepackage{amsmath}
\usepackage{amssymb}
\usepackage{booktabs}
\usepackage{times}
\usepackage{epsfig}
\usepackage{graphicx}
\usepackage{amsmath}
\usepackage{amssymb}
\usepackage{algorithm}
\usepackage{algorithmicx}
\usepackage{algpseudocode}
\usepackage[usestackEOL]{stackengine}
\usepackage{adjustbox}
\usepackage{multirow}
\usepackage{mathtools}
\usepackage{tabularx}
\usepackage{xcolor}

\DeclarePairedDelimiter\floor{\lfloor}{\rfloor}

\newcommand\eatpunct[1]{}

\newcolumntype{Y}{>{\centering\arraybackslash}X}

\usepackage{flushend}

\usepackage[pagebackref,breaklinks,colorlinks]{hyperref}

\usepackage[capitalize]{cleveref}
\crefname{section}{Sec.}{Secs.}
\Crefname{section}{Section}{Sections}
\Crefname{table}{Table}{Tables}
\crefname{table}{Tab.}{Tabs.}

\def\cvprPaperID{2366} %
\def\confName{CVPR}
\def\confYear{2022}

\input{preamble}
\begin{document}
\title{Multiview Transformers for Video Recognition}

\author{Shen Yan$^\ddag$\thanks{
This work was done while the first author was an intern at Google.} \;\; Xuehan Xiong$^\dag$ \;\; Anurag Arnab$^\dag$ \;\; Zhichao Lu$^\dag$ \;\; Mi Zhang$^\ddag$\;\; \\
Chen Sun$^\dag$$^\S$\;\; Cordelia Schmid$^\dag$ \\
$^\dag$Google Research \quad
$^\ddag$Michigan State University \quad
$^\S$Brown University \\
{\tt \small \{xxman, aarnab, lzc, chensun, cordelias\}@google.com \;\; \{yanshen6, mizhang\}@msu.edu \;\;}\\
}

\maketitle
\begin{abstract}

\input{abstract}

\end{abstract}

\input{intro}

\input{related}

\input{method}

\input{results}

\input{conc}

{\small
\bibliographystyle{ieee_fullname}
\bibliography{main}
}

\newpage

\input{supplementary}

\clearpage

\end{document}

%% file: preamble.tex
\usepackage{overpic}
\usepackage{enumitem} %
\usepackage{overpic} %
\usepackage{color}

\definecolor{turquoise}{cmyk}{0.65,0,0.1,0.3}
\definecolor{purple}{rgb}{0.65,0,0.65}
\definecolor{dark_green}{rgb}{0, 0.5, 0}
\definecolor{orange}{rgb}{0.8, 0.6, 0.2}
\definecolor{red}{rgb}{0.8, 0.2, 0.2}
\definecolor{darkred}{rgb}{0.6, 0.1, 0.05}
\definecolor{blueish}{rgb}{0.0, 0.3, .6}
\definecolor{light_gray}{rgb}{0.7, 0.7, .7}
\definecolor{pink}{rgb}{1, 0, 1}
\definecolor{greyblue}{rgb}{0.25, 0.25, 1}

\usepackage{blindtext}

\renewcommand{\paragraph}[1]{\vspace{1em}\noindent\textbf{#1}.}

%% file: abstract.tex
Video understanding requires reasoning at multiple spatiotemporal resolutions -- from short fine-grained motions to events taking place over longer durations.
Although transformer architectures have recently advanced the state-of-the-art, they have not explicitly modelled different spatiotemporal resolutions.
To this end, we present Multiview Transformers for Video Recognition (MTV).
Our model consists of separate encoders to represent different views of the input video with lateral connections to fuse information across views. 
We present thorough ablation studies of our model and show that MTV consistently performs better than single-view counterparts in terms of accuracy and computational cost across a range of model sizes.
Furthermore, we achieve state-of-the-art results on six standard datasets, and improve even further with large-scale pretraining.
Code and checkpoints are available at: \href{https://github.com/google-research/scenic/tree/main/scenic/projects/mtv}{https://github.com/google-research/scenic}. %

%% file: intro.tex
\section{Introduction} 

Vision architectures based on convolutional neural networks (CNNs), and now more recently transformers, have made great advances in numerous computer vision tasks.
A central idea, that has remained constant across classical methods based on handcrafted features~\cite{burt1987laplacian, dalal2005histograms, lazebnik2006beyond} to CNNs~\cite{liu2016ssd,lin2017feature, zhao2017pyramid} and now transformers~\cite{wang2021pyramid, chen2021crossvit, liu2021swin}, has been to analyze input signals at multiple resolutions.

In the image domain, multiscale processing is typically performed with pyramids as the statistics of natural images are isotropic (all orientations are equally likely) and shift invariant~\cite{huang1999statistics, torralba2003statistics}.
To model multiscale temporal information in videos, previous approaches such as SlowFast~\cite{feichtenhofer_iccv_2019} have processed videos with two streams, using a ``Fast'' stream operating at high frame rates and a ``Slow'' stream at low frame rates, or employed graph neural networks to model long-range interactions~\cite{wu2019long,arnab_graph_structured_iccv_2021}.

\input{teaser.tex}

When creating a pyramidal structure, spatio-temporal information is partially lost due to its pooling or subsampling operations.
For example, when constructing the ``Slow'' stream, SlowFast~\cite{feichtenhofer_iccv_2019} subsamples frames, losing temporal information.
In this work, we propose a simple transformer-based model without relying on pyramidal structures or subsampling the inputs to capture multi-resolution temporal context.
We do so by leveraging multiple input representations, or ``views'' of the input video.
As shown in Fig.~\ref{fig:model-overview}, we extract tokens from the input video over multiple temporal durations. %
Intuitively, tokens extracted from long time intervals capture the gist of the scene (such as the background where the activity is taking place), whilst tokens extracted from short segments can capture fine-grained details %
(such as the gestures performed by a person).

We propose a multiview transformer (Fig.~\ref{fig:model-overview}) to process these tokens, and it consists of separate transformer encoders specialized for each ``view'', with lateral connections between them to fuse information from different views to each other.
We can use transformer encoders of varying sizes to process each view, and find that it is better (in terms of accuracy/computation trade-offs) to use a smaller encoder (\eg smaller hidden sizes and fewer layers) to represent the broader view of the video (Fig.~\ref{fig:model-overview} left) while an encoder with larger capacity is used to capture the details (Fig.~\ref{fig:model-overview} right). %
This design therefore poses a clear contrast to pyramid-based approaches where model complexity increases as the spatio-temporal resolution decreases.  %
Our design is verified by our experiments which show clear advantages over the former approach.

Our proposed method, of processing different ``views'' of the input video is simple, and in contrast to previous work~\cite{feichtenhofer_iccv_2019} generalizes readily to a variable number of views.
This is significant, as our experiments show that accuracy increases as the number of views grows.
Although our proposed architecture increases the number of tokens processed by the network according to the number of input views, we show that we can consistently achieve superior accuracy/computation trade-offs compared to the current state of the art~\cite{arnab2021vivit}, across a spectrum of model sizes, ranging from ``Small'' to ``Huge''.
We show empirically that this is because processing more views in parallel enables us to achieve larger accuracy improvements than increasing the depth of the transformer network. 
We perform thorough ablation studies of our design choices, and achieve state-of-the-art results on six 
standard video classification datasets. Moreover, we show that these results can be further improved with large-scale pretraining.

%% file: teaser.tex
\begin{figure}[t]
    \vspace{-\baselineskip}
	\centering
    \includegraphics[width=0.9\columnwidth]{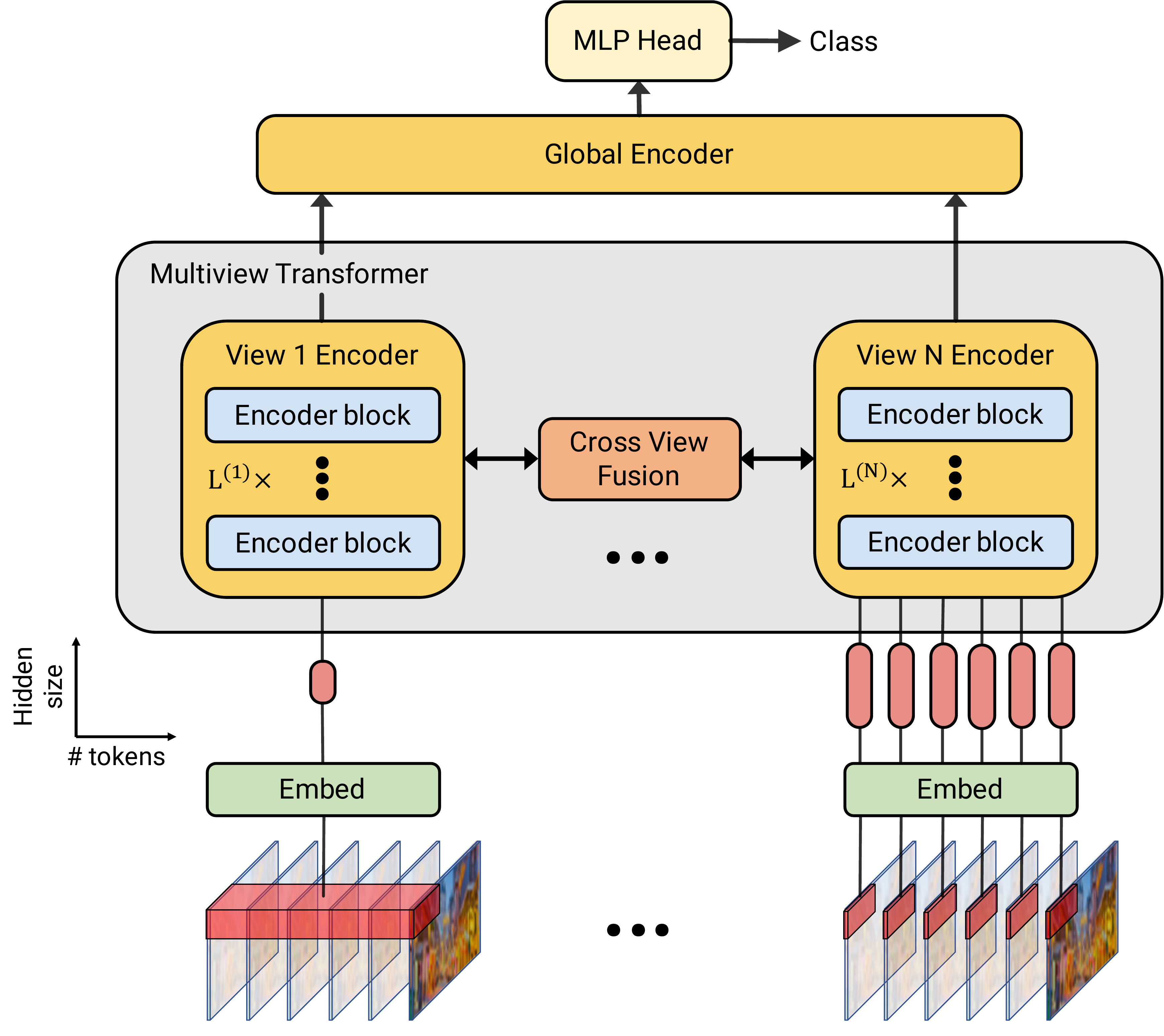}
    \caption{
        Overview of our Multiview Transformer.
    We create multiple input representations, or ``views'', of the input, by tokenizing the video using tubelets of different sizes (for clarity, we show two views here).
    These tokens are then processed by separate encoder streams, which include lateral connections and a final global encoder to fuse information from different views.
    Note that the tokens from each view may have different hidden sizes, and the encoders used to process them can vary in architecture too.
    }
    \vspace{-\baselineskip}
	\label{fig:model-overview}
\end{figure}

%% file: related.tex
\section{Related Work} 
\label{sec:related}

\paragraph{Evolution of video understanding models} Early works~\cite{laptev2005space,klaser2008spatio,wang2013dense} relied on hand-crafted features to encode motion and appearance information. With the emergence of large labelled datasets like ImageNet~\cite{deng2009imagenet}, Convolutional Neural Networks (CNNs)~\cite{lecun1989backpropagation} showed their superiority over the classic methods. Since AlexNet~\cite{krizhevsky_neurips_2012} won the ImageNet challenge by a large margin, CNNs have been quickly adopted to various vision tasks, %
their architectures have been refined over many generations~\cite{simonyan2014very,szegedy2015going,he2016deep,chollet2017xception} and later improved by Neural Architecture Search (NAS)~\cite{real2017large,zoph2016neural,tan2019efficientnet}.
At the same time, CNNs and RNNs have quickly become the de-facto backbones for video understanding tasks~\cite{karpathy2014large,simonyan_neurips_2014,ng_cvpr_2015}. Since the release of the Kinetics dataset~\cite{kay_arxiv_2017}, 3D CNNs~\cite{tran_iccv_2015,feichtenhofer_neurips_2016,Carreira_2017_CVPR} have gained popularity, and many variants~\cite{sun_iccv_2015,tran_cvpr_2018,xie_s3d_eccv_2018,tran_iccv_2019,feichtenhofer_cvpr_2020} have been developed to improve the speed and accuracy.
Convolution operations can only process one local neighborhood at a time, and consequently, transformer blocks~\cite{vaswani2017attention} have been inserted into CNNs as additional layers to improve modeling of long range interactions among spatio-temporal features~\cite{wang2018nonlocal,wang_nas_eccv_2020}.
Although achieving great success in natural language~\cite{devlin_naacl_2019, brown2020language, raffel2019exploring}, pure transformer architectures had not gained the same popularity in computer vision until Vision Transformers (ViT)~\cite{dosovitskiy2020image}.
Inspired by ViT, ViViT~\cite{arnab2021vivit} and Timesformer~\cite{bertasius_arxiv_2021} were the first two works that successfully adopted a pure transformer architecture for video classification, advancing the state of the art previously set by 3D CNNs.

\paragraph{Multiscale processing in computer vision}
``Pyramid’’ structures~\cite{adelson1984pyramid} are one of the most popular multiscale representations for images and have been key in the early computer vision works, where  their use has been widespread in multiple domains including 
feature descriptors~\cite{lowe2004distinctive}, feature tracking~\cite{lucas1981iterative,bouguet2001pyramidal}, image compression~\cite{burt1987laplacian}, etc. This idea has also been successfully adopted for modern CNNs~\cite{simonyan2014very,szegedy2015going,he2016deep} where the spatial dimension of the network is gradually reduced while the network ``depth’’ is gradually increased to encode more semantically rich features. Also, this technique has been used to produce  higher resolution output features for downstream tasks\cite{lin2017feature, zhao2017pyramid, liu2016ssd}. Multiscale processing is necessary for CNNs because a convolution operation only operates on a sub-region of the input and a hierarchical structure is required to capture the whole view of the image or video. In theory, such a hierarchy is not required for transformers as each token ``attends’’ to all other positions. In practice, due to the limited amount of training data, applying similar multiscale processing in transformers~\cite{fan2021multiscale,liu2021swin,chen2021crossvit,wang2021pyramid} to reduce complexity of the model has proven to be effective.

Our model does not follow the pyramid structure but directly takes different views of the video and feeds them into cross-view encoders. As our experiments validate, this alternative multiview architecture has consistently outperformed its single-view counterpart in terms of accuracy/FLOP trade-offs.
This is because processing more views in parallel gives us larger accuracy improvements than increasing the depth of the transformer network.
Significantly, such improvement persists as we scale the model capacity to over a billion parameters (\eg, our ``Huge'' model), which has not been shown by the previous pyramid-structured transformers~\cite{fan2021multiscale,liu2021swin,wang2021pyramid}.
Conceptually, our method is most comparable to SlowFast~\cite{feichtenhofer_iccv_2019} where a two-stream CNN is used to process two views of the same video clip (densely sampled and sparsely sampled frames).
Instead of sampling the input video at different frame rates, we obtain different view by linearly projecting spatio-temporal ``tubelets''~\cite{arnab2021vivit} of varying sizes for each view.
Furthermore, we empirically show that our proposed method outperforms ~\cite{feichtenhofer_iccv_2019} when using transformer backbones.

%% file: method.tex
\section{Multiview Transformers for Video}

We begin with an overview of vision transformer, ViT~\cite{dosovitskiy2020image}, and its extension to video, ViViT~\cite{arnab2021vivit}, which our model is based on, in Sec.~\ref{sec:method_prelim}.
As shown in Fig.~\ref{fig:model-overview}, our model constructs different ``views'' of the input video by extracting tokens from spatio-temporal tubelets of varying dimensions (Sec.~\ref{sec:method_multiview_tokenization}).
These tokens are then processed by a multiview transformer, which incorporates lateral connections to efficiently fuse together information from multiple scales (Sec.~\ref{sec:method_multiview_transformer}).

\subsection{Preliminaries: ViT and ViViT}
\label{sec:method_prelim}

We denote our input video as $\mathbf{V} \in \mathbb{R}^{T \times H \times W \times C}$.
Transformer architectures~\cite{vaswani2017attention} process inputs by converting inputs into discrete tokens which are subsequently processed by multiple transformer layers sequentially.

ViT~\cite{dosovitskiy2020image} extracts tokens from images by partitioning an image into non-overlapping patches and linearly projecting them.
ViViT~\cite{arnab_graph_structured_iccv_2021} extends this to video by extracting $N$ non-overlapping, spatio-temporal ``tubes''\cite{arnab2021vivit} from the input video, $x_1, x_2, \ldots x_N \in \mathbb{R}^{t \times h \times w \times c}$ where $N = \floor{\frac{T}{t}} \times \floor{\frac{H}{h}} \times \floor{\frac{W}{w}}$. 

Each tube, $x_i$, is then projected into a token, $\mathbf{z}_i \in \mathbb{R}^d$ by a linear operator $\mathbf{E}$, as $\mathbf{z}_i = \mathbf{E}x_i$.
All tokens are then concatenated together to form a sequence, which is prepended with a learnable class token $\mathbf{z}_{cls} \in \mathbb{R}^{d}$~\cite{devlin_naacl_2019}.
As transformers are permutation invariant, a positional embedding $\mathbf{p} \in \mathbb{R}^{(N + 1) \times d}$, is also added to this sequence.
Therefore, this tokenization process can be denoted as
\begin{equation}
    \mathbf{z}^{0} = [\mathbf{z}_{cls}, \mathbf{E}x_1, \mathbf{E}x_2, \ldots, \mathbf{E}x_N] + \mathbf{p}.
\end{equation}
Note that the linear projection $\mathbf{E}$ can also be seen as a 3D convolution with a kernel of size $t \times h \times w$ and stride of $(t, h, w)$ in the time, height and width dimensions respectively.

The sequence of tokens $\mathbf{z}$ is then processed by a transformer encoder consisting of $L$ layers.
Each layer, $\ell$, is applied sequentially, and consists of the following operations,
\begin{align}
 \mathbf{y}^{\ell} &= \text{MSA}\left(\text{LN}\left(\mathbf{z}^{\ell - 1}\right)\right) + \mathbf{z}^{\ell - 1}, \label{eq:transformer_layer_msa} \\
 \mathbf{z}^{\ell} &= \text{MLP}\left(\text{LN}\left(\mathbf{y}^{\ell}\right) \right) + \mathbf{y}^{\ell}
 \label{eq:transformer_layer_mlp}
\end{align}
where $\text{MSA}$ denotes multi-head self-attention~\cite{vaswani2017attention}, $\text{LN}$ is layer normalization~\cite{ba2016layer} and $\text{MLP}$ consists of two linear projections separated by GeLU~\cite{hendrycks2016gaussian} non-linearity.

Finally, a linear classifier, $\mathbf{W}^{\text{out}} \in \mathbb{R}^{d \times C}$ maps the encoded classification token, $\textbf{z}^{\ell}_{cls}$ to one of $C$ classes. 

\input{cross_view_fusion}

\subsection{Multiview tokenization}
\label{sec:method_multiview_tokenization}

In our model, we extract multiple sets of tokens, $\mathbf{z}^{0, (1)}, \mathbf{z}^{0, (2)}, \ldots, \mathbf{z}^{0, (V)}$ from the input video.
Here, $V$ is the number of views, and thus $\mathbf{z}^{\ell, (i)}$ denotes tokens after $\ell$ layers of transformer processing for the $i^{th}$ view. We define a view as a video representation expressed by a set of fixed-sized tubelets.
A larger view corresponds to a set of larger tubelets (and thus fewer tokens) and a smaller view corresponds to smaller tubelets (and thus more tokens).
The $0^{th}$ layer corresponds to the tokens that are input to the subsequent transformer.
As shown in Fig.~\ref{fig:model-overview}, we tokenize each view using a 3D convolution, as it was the best tokenization method reported by~\cite{arnab2021vivit}.
We can use different convolutional kernels, and different hidden sizes, $d^{(i)}$, for each view.
Note that smaller convolutional kernels correspond to smaller spatio-temporal ``tubelets'', thus resulting in more tokens to be processed for the $i^{th}$ view. 
Intuitively, fine-grained motions can be captured by smaller tubelets whilst larger tubelets capture slowly-varying semantics of the scene. %
As each view captures different levels of information, we use  transformer encoders of varying capacities for each stream with lateral connections between them to fuse information, as described in the next section.

\subsection{Multiview transformer}
\label{sec:method_multiview_transformer}

After extracting tokens from multiple views, we have $\mathbf{Z}^{0} = [\mathbf{z}^{0, (1)}, \mathbf{z}^{0, (2)}, \ldots, \mathbf{z}^{0, (V)}]$
from the input, which are processed with a multiview transformer as shown in Fig.~\ref{fig:model-overview}. As self-attention has quadratic complexity~\cite{vaswani2017attention}, processing tokens from all views jointly is not computationally feasible for video.
As a result, we first use a multiview encoder, comprising of separate transformer encoders (consisting of $L^{(i)}$ transformer layers) for the tokens between views, with lateral connections between these encoders to fuse information from each view (Fig.~\ref{fig:three variants}).
Finally, we extract a token representation from each view, and process these jointly with a final global encoder to produce the final classification token, which we linearly read-off to obtain the final classification.

 \vspace{-0.1\baselineskip}
\subsubsection{Multiview encoder}
\label{sec:method_multiview_encoder}

Our multiview encoder consists of separate transformer encoders for each view which are connected by lateral connections to fuse cross-view information.
Each transformer layer within the encoders follows the same design as the original transformer of Vaswani~\etal~\cite{vaswani2017attention}, except for the fact that we optionally fuse information from other streams within the layer as described in Sec.~\ref{sec:cross_view_fusion}.
Note that our model is agnostic to the exact type of transformer layer used.
Furthermore, within each transformer layer, we compute self-attention only among  tokens extracted from the same temporal index, following the Factorised Encoder of~\cite{arnab2021vivit}.
This significantly reduces the computational cost of the model.
Furthermore, self-attention along all spatio-temporal tokens is unnecessary, as we fuse information from other views within the multiview encoder, and also because of the subsequent global encoder which aggregates tokens from all streams.

\subsubsection{Cross-view fusion}
\label{sec:cross_view_fusion}

We consider the following three cross-view fusion methods.
Note that the hidden dimensions of the tokens, $d^{(i)}$, can vary between  views.

\paragraph{Cross-view attention (CVA)}
A straight-forward method of combining information between different views is to perform self-attention jointly on all $\sum_i N^{(i)}$ tokens where $N^{(i)}$ is the number of tokens in the $i^{th}$ view.
However, due to the quadratic complexity of self-attention, this is prohibitive computationally for video models, and hence we perform a more efficient alternative.

We sequentially fuse information between all pairs of two adjacent views, $i$ and $i + 1$, where the views are ordered in terms of increasing numbers of tokens (\ie $N^{(i)} \leq N^{(i + 1)}$).
Concretely, to update the tokens from the larger view, $\mathbf{z}^{(i)}$, we compute attention where the queries are $\mathbf{z}^{(i)}$, and the keys and values are $\mathbf{z}^{(i+1)}$ (the tokens from the smaller view).
As the hidden dimensions of the tokens between the two views can be different, we first project the keys and values to the same dimension, as denoted by
\begin{align}
    \mathbf{z}^{(i)} &= \text{CVA}(\mathbf{z}^{(i)}, \mathbf{W}^{\text{proj}}
    \mathbf{z}^{(i+1)}), \\
    \text{CVA}(\mathbf{x}, \mathbf{y}) &= \text{Softmax}\left(\frac{\mathbf{W}^{Q}\mathbf{x} \mathbf{W}^{K}\mathbf{y}^{\top}}{\sqrt{d_k}}\right) \mathbf{W}^{V}\mathbf{y}.
\end{align}
Note that $\mathbf{W}^{Q}$, $\mathbf{W}^{K}$ and $\mathbf{W}^{V}$ are the query-, key- and value-projection matrices used in the attention operation~\cite{vaswani2017attention}.
As shown in Fig.~\ref{fig:cva}, we also include a residual connection around the cross-view attention operation, and zero-initialize the parameters of this operation, as this helps when using image-pretrained models as is common practice~\cite{arnab2021vivit, bertasius_arxiv_2021}. Similar studies on cross stream attention have been done by~\cite{chen2021crossvit} for images.

\paragraph{Bottleneck tokens} 
An efficient method of transferring information between tokens from two views, $\mathbf{z}^{(i)}$ and $\mathbf{z}^{(i+1)}$, is by an intermediate set of $B$ bottleneck tokens.
Once again, we sequentially fuse information between all pairs of two adjacent views, $i+1$ and $i$, where the views are ordered in terms of increasing numbers of tokens.

In more detail, we initialize a sequence of bottleneck tokens, $\mathbf{z}_B^{(i+1)} \in \mathbb{R}^{B^{(i+1)} \times d^{(i+1)}}$ where $B^{(i+1)}$ is the number of bottleneck tokens in the $(i+1)^{th}$ view and $B^{(i+1)} \ll N^{(i+1)}$.
As shown in Fig.~\ref{fig:ib} (where $B = 1$), the bottleneck tokens from view $i+1$, $\mathbf{z}_{B}^{(i+1)}$, are concatenated to the input tokens of the same view, $\mathbf{z}^{(i+1)}$, and processed with self-attention.
This effectively transfers information between all tokens from view $i+1$.
Thereafter, these tokens, $\mathbf{z}_{B}^{(i+1)}$ are linearly projected to the depth of view $i$, and concatenated to $\mathbf{z}^{(i)}$ before performing self-attention again.
This process is repeated between each pair of adjacent views as shown in Fig.~\ref{fig:ib}, and allows us to efficiently transfer information from one view to the next.

As with cross-view attention, we sequentially perform fusion between all pairs of adjacent views, beginning from the view with the largest number of tokens, and proceeding in order of decreasing token numbers.
Intuitively, this allows the view with the fewest tokens to aggregate fine-grained information from all subsequent views.

Note that the only parameters introduced into the model from this fusion method are the linear projections of bottleneck tokens from one view to the next, and the bottleneck tokens themselves which are learned from random initialization.
We also note that ``bottleneck'' tokens have also been used by~\cite{jaegle2021perceiver, nagrani2021attention}.

\paragraph{MLP fusion}
Recall that each transformer encoder layer consists of a multi-head self attention operation (Eq.~\ref{eq:transformer_layer_msa}), followed by an MLP block (Eq.~\ref{eq:transformer_layer_mlp}).
A simple method is to fuse before the MLP block within each encoder layer.

Concretely, as shown in Fig.~\ref{fig:mlp}, tokens from view $i+1$, $\mathbf{z}^{(i+1)}$ with hidden dimension $d^{(i+1)}$ are concatenated with tokens from view $i$ along the hidden dimension.
These tokens are then fed into the MLP block of layer $i$ and linearly projected to the depth $d^{(i)}$.
This process is repeated between adjacent views of the network, where once again, views are ordered by increasing number of tokens per view.

\paragraph{Fusion locations}
We note that it is not necessary to perform cross-view fusion at each layer of the cross-view encoder to transfer information among the different views, since each fusion operation has a global ``receptive field'' that considers all the tokens from the previous views.
Furthermore, it is also possible for the encoders for each individual view to have different depths, meaning that fusion can occur between layer $l$ of view $i$ and layer $l'$ of view $j$ where $l \neq l'$.
Therefore, we consider the fusion locations as a design choice which we perform ablation studies on.

\subsubsection{Global encoder}
\label{sec:method_global_encoder}

Finally, we aggregate the tokens from each of the views with the final global encoder, as shown in Fig.~\ref{fig:model-overview}, effectively fusing information from all views after the cross-view transformer.
We extract the classification token from each view, $\{\mathbf{z}_{cls}^{(i)}\}_{i=1}^{V}$, and process them further with another transformer encoder, following Vaswani~\etal~\cite{vaswani2017attention}, that aggregates information from all views.
The resulting classification token is then mapped to one of $C$ classification outputs, where $C$ is the number of classes.

%% file: cross_view_fusion.tex
\begin{figure*}
     \vspace{-0.5\baselineskip}
     \centering
     \begin{subfigure}[b]{0.3\textwidth}
         \centering
         \includegraphics[width=\textwidth]{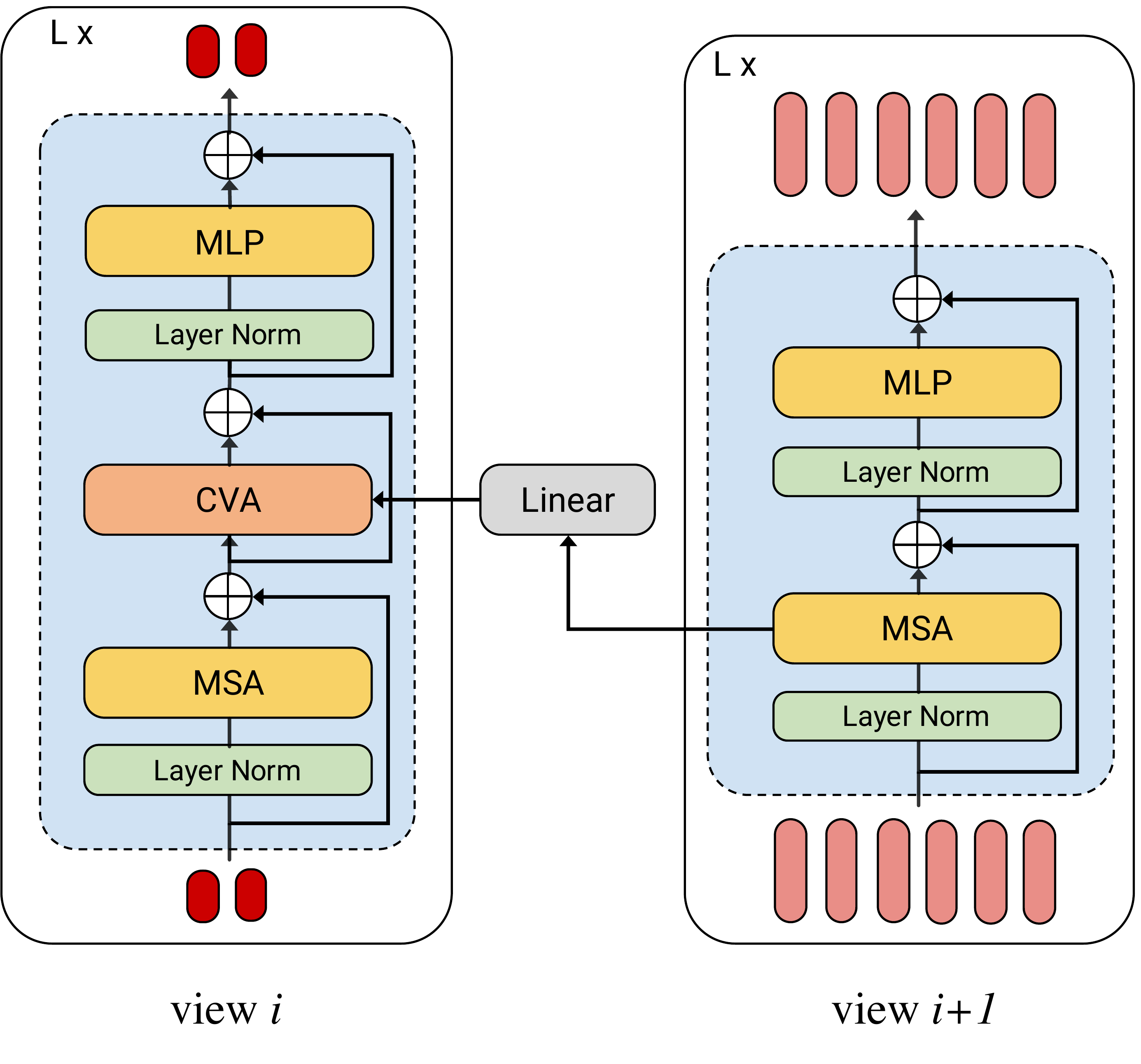}
         \caption{An example of CVA for fusion.}
         \label{fig:cva}
     \end{subfigure}
     \hfill
     \begin{subfigure}[b]{0.3\textwidth}
         \centering
         \includegraphics[width=\textwidth]{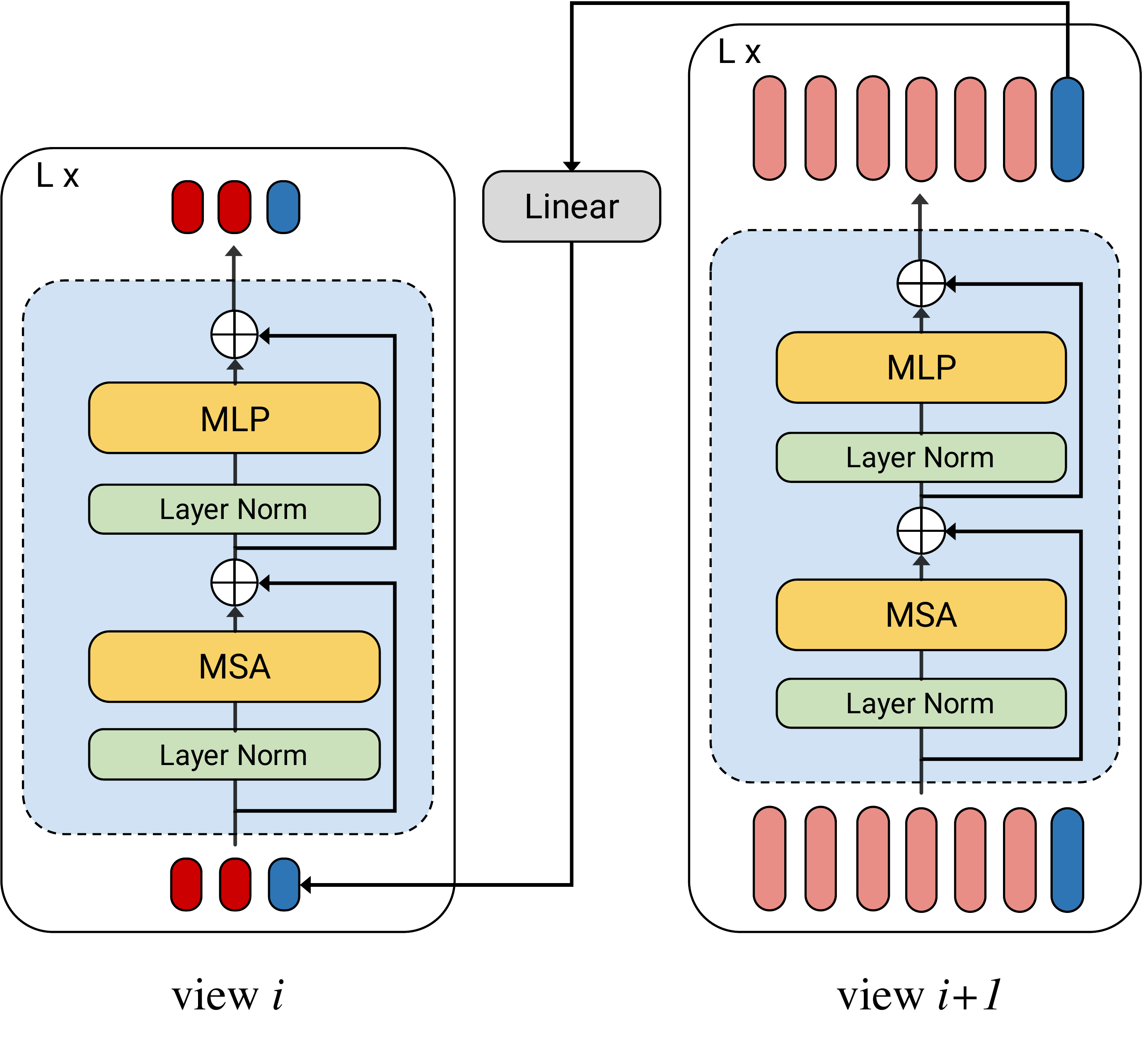}
         \caption{An example of bottleneck tokens for fusion.}
         \label{fig:ib}
     \end{subfigure}
     \hfill
     \begin{subfigure}[b]{0.3\textwidth}
         \centering
         
         \includegraphics[width=\textwidth]{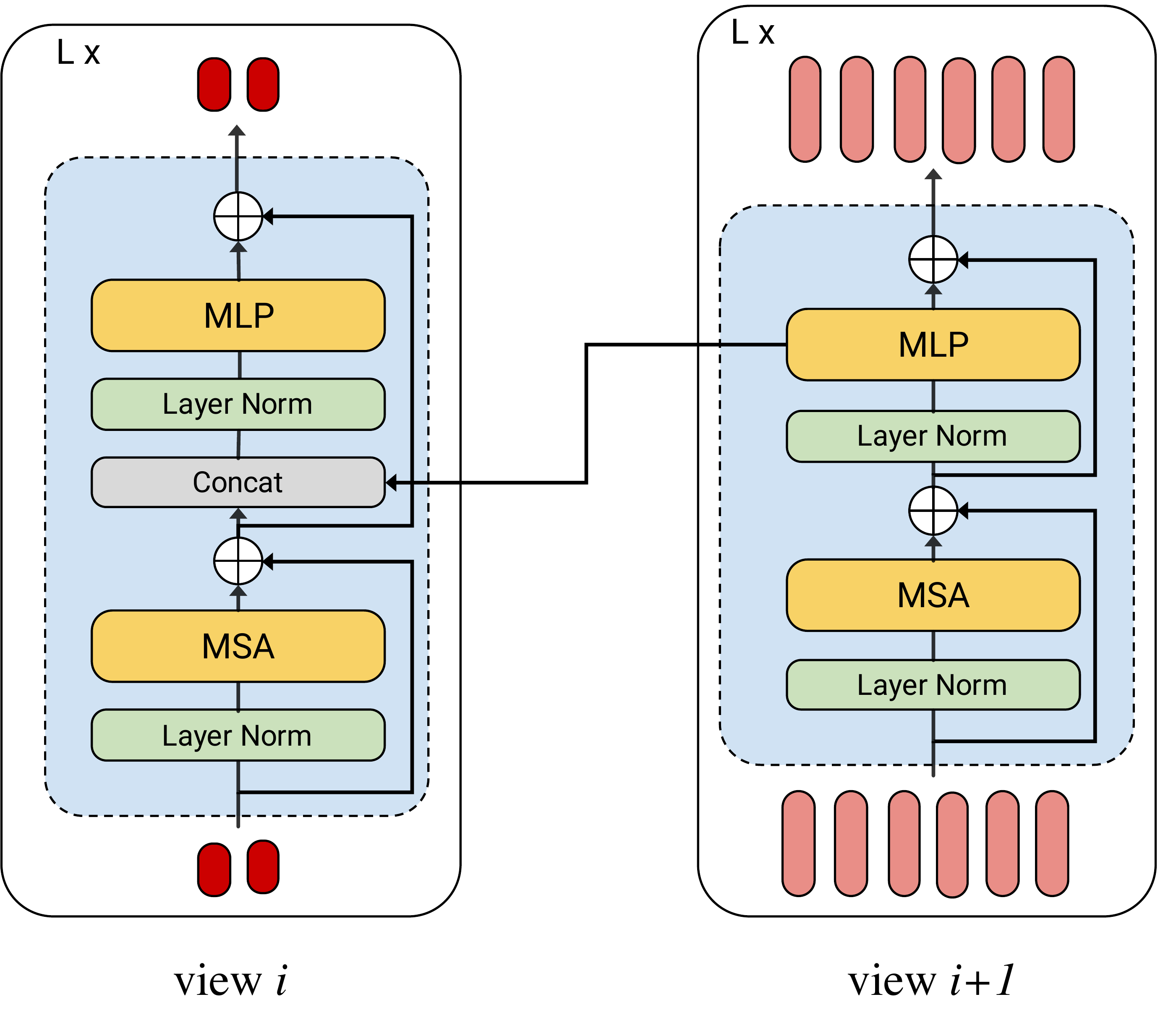}
        \caption{An example of MLP fusion.}
         \label{fig:mlp}
     \end{subfigure}
        \caption{An illustration of our proposed cross-view fusion methods. In all three subfigures, view $i$ (left) refers to a video representation using larger tubelets, and thus less input tokens and view $i+1$ (right) corresponds to the representation with smaller tubelets and more input tokens. ``$+$’’ denotes summation. Tokens extracted from tubelets are colored red and bottleneck tokens are colored blue. MSA is short for Multihead Self-Attention and CVA stands for Cross View Attention.
        }
        \label{fig:three variants}
    \label{sec: cross-view attn}
    \vspace{-\baselineskip}
\end{figure*}

%% file: results.tex
\section{Experiments}
\subsection{Experimental setup}

\paragraph{Model variants} For the backbone of each view, we consider five ViT variants, ``Tiny'', ``Small'', ``Base'', ``Large'', and ``Huge''. Their settings strictly follow the ones defined in BERT~\cite{devlin_naacl_2019} and ViT~\cite{dosovitskiy2020image, steiner2021train}, \ie number of transformer layers, number of attention heads, hidden dimensions.
See Appendix \ref{sec: model_configs} for the detailed settings.
For convenience, each model variant is denoted with the following abbreviations indicating the backbone size and tubelet length.
For example, B/2+S/4+Ti/8 denotes a three-view model, where a ``Base'', ``Small'', and ``Tiny'' encoders are used to processes tokens from the views with tubelets of sizes $16\times16\times2$, $16\times16\times4$, and $16\times16\times8$, respectively.
Note that we omit 16 in our model abbreviations because all our models use $16\times16$ as the spatial tubelet size except for the ``Huge'' model, which uses $14\times14$, following ViT~\cite{dosovitskiy2020image}.
All model variants use the same global encoder which follows the ``Base'' architecture, except that the number of heads is set to 8 instead of 12.
The reason is that the hidden dimension of the tokens should be divisible by the number of heads for multi-head attention, and the number of hidden dimensions across all standard transformer architectures (from ``Tiny'' to ``Huge''~\cite{steiner2021train, dosovitskiy2020image}) is divisible by 8.

\input{ablation}
\paragraph{Training and inference} 
We follow the training settings of ViViT reported in the paper and public code~\cite{arnab2021vivit}, unless otherwise stated. 
Namely, all models are trained on 32 frames with a temporal stride of 2.
We train our model using synchronous SGD with momentum of 0.9 following a cosine learning rate schedule with a linear warm up.
The input frame resolution is set to be $224\times224$ in both training and inference. We follow~\cite{arnab2021vivit} and apply the same data augmentation and regularization schemes~\cite{huang_stochasticdepth_eccv_2016,zhang_mixup_iclr_2018,cubuk_arxiv_2019,szegedy_cvpr_2016}, which were used by~\cite{touvron2021training} to train vision transformers more effectively.
During inference, we adopt the standard evaluation protocol by averaging over multiple spatial and temporal crops.
The number of crops is given in the results tables.
For reproducibility, we include exhaustive details in Appendix \ref{sec: hparams}.

\paragraph{Initialization}
Following previous works~\cite{arnab2021vivit, bertasius_arxiv_2021,patrick2021keeping}, we initialize our model from a corresponding ViT model pretrained on large-scale image datasets~\cite{deng2009imagenet, 300m_iccv17} obtained from the public code of~\cite{dosovitskiy2020image}.
The initial tubelet embedding operator, $\mathbf{E}$, and positional embeddings, $\mathbf{p}$, have different shapes in the pretrained model and we use the same technique as~\cite{arnab2021vivit} to adapt them to initialize each view of our multiview encoder (Sec.~\ref{sec:method_multiview_encoder}).
The final global encoder (Sec.~\ref{sec:method_global_encoder}) is randomly initialized.

\paragraph{Datasets} We report the performance of our proposed models on a diverse set of video classification datasets:

\emph{Kinetics}~\cite{kay_arxiv_2017} is a collection of large-scale, high-quality datasets of 10s video clips focusing on human actions. We report results on Kinetics 400, 600, and 700, with 400, 600, and 700 classes, respectively.

\emph{Moments in Time}~\cite{monfort_pami_2019} is a collection of 800,000 labeled 3~second videos, involving people, animals, objects or natural phenomena, that capture the gist of a dynamic scene.

\emph{Epic-Kitchens-100}~\cite{damen2021rescaling} consists of 90,000 egocentric videos, totaling 100 hours, recorded in kitchens. %
Each video is labeled with a ``noun'' and a ``verb'' and therefore we predict both categories using a single network with two ``heads''. Three accuracy scores (``noun'', ``verb'', and ``action'') are commonly reported for this dataset with action accuracy being the primary metric. The “action” label is formed by selecting the top-scoring noun and verb pair. 

\emph{Something-Something V2}~\cite{goyal_iccv_2017} consists of more than 220,000 short video clips that show humans interacting with everyday objects. Similar objects and backgrounds appear in videos across different classes. Therefore, in contrast to other datasets, this one challenges a model’s capability to distinguish classes from motion cues.

\subsection{Ablation study} \label{sec:ablations}

We conduct ablation studies on the Kinetics 400 dataset.
In all cases, the largest backbone in the multiview encoder is ``Base'' for faster experimentation.
We report accuracies when averaging predictions across multiple spatio-temporal crops, as standard practice~\cite{arnab2021vivit, bertasius_arxiv_2021, feichtenhofer_iccv_2019,Carreira_2017_CVPR}.
In particular, we use $4 \times 3$ crops, that is 4 temporal crops, with 3 spatial crops for each temporal crop.
We used a learning rate of 0.1 for all experiments for 30 epochs, and used no additional regularization as done by~\cite{arnab2021vivit}.

\input{vivit_vs_mtv}

\paragraph{Model-view assignments}
Recall that a view is a video representation in terms of tubelets, and that a larger view equates to larger tubelets (and hence fewer transformer tokens) and smaller views correspond to smaller tubelets (and thus more tokens).

We considered two model-view assignment strategies: larger models for larger views (\eg, B/8+Ti/2, the larger ``Base'' model is used to encode $16\times 16 \times 8$ tubelets and the smaller ``Tiny'' model encodes $16\times 16 \times 2$ tubelets) and smaller models for larger views (\eg, B/2+Ti/8).
Table~\ref{tab:model_view_assignment} shows that assigning a larger model to smaller views is superior. For example, B/2+S/4+Ti/8 scores 81.8\% while B/8+S/4+Ti/2 only scores 78.5\%. One may argue that this is due to the increase of the FLOPs but B/4+S/8+Ti/16 still outperforms B/8+S/4+Ti/2 by a large margin under similar FLOPs. Our explanation is that larger views capture the gist of the scene, which requires less complexity to learn while the details of the scene are encapsulated by smaller views so a larger-capacity model is needed.

Another strategy is to assign the same model to all views. Table~\ref{tab:same_model_different_views} shows that in all three examples there is little difference between assigning a ``Base'' model and assigning a ``Small'' or ``Tiny'' model to larger views.
This result is surprising yet beneficial since we can reduce the complexity of the model at almost no cost of accuracy.

\paragraph{What is the best cross-view fusion method?\eatpunct}
Table~\ref{tab:fusion_methods} shows the comparison of different fusion methods on a three-view model.
We use one late fusion and an ensemble approach as the baselines.
``Ensemble'' simply sums the probabilities produced from each view, where the models from each view are trained separately. We also tried summing up the logits and majority voting but both obtained worse results. This method actually decreases the performance compared to the B/4 model since ``Small'' and ``Tiny'' models perform not comparably well. %
``Late fusion'' concatenates the final embeddings produced by the transformer encoder from each view without any cross-view operations before feeding it into the global encoder. It improves the B/4 model from 78.3\% to 80.6\%.
All of our fusion methods except MLP outperform the baselines while CVA is the best overall.
Based on this observation, we choose CVA as the fusion method for all subsequent experiments. %
MLP fusion is the worst performing method of the three and we think it is because concatenation in the MLP blocks introduces additional channels that have to be randomly initialized, making model optimization more difficult. %

\paragraph{Effect of the number of views}
Table~\ref{tab:num_views} shows performance on Kinetics-400 as we increase the number of views.
With two views we achieve a \textbf{+2.5}\% in Top-1 accuracy over the baseline B/4 model.
As we increase to three views, the improvement widens to \textbf{2.8}\%.
Furthermore, we show that such improvement is non-trivial. For example, we also train a 14-layer and a 17-layer variants of the ``Base'' model. They share similar FLOPs with our two-view and three-view counterparts but their performance remains similar to that of the baseline.

\paragraph{Which layers to apply cross-view fusion?}
Motivated by Tab.~\ref{tab:fusion_methods}, we fix the fusion method to CVA, and vary the locations and number of layers where we apply CVA, when using a three-view B+S+Ti model (each encoder thus has 12 layers) in Tab.~\ref{tab:cva_layers}.
The choices are in the early-, mid-, and late-stages of the transformer encoders and the number of fusion layers is set to be one and two.
When using one fusion layer, the best location for fusion is mid followed by late, then early.
Adding more fusion layers in the same stage does not improve the performance but combining mid and late fusion improves the performance.
For example, fusion at 5th and 11th layers achieve the best result.
Based on this observation, we set the fusion layers to be \{11, 23\} for L+B+S+Ti and \{11, 23, 31\} for H+B+S+Ti model variants, respectively, in subsequent experiments.

\paragraph{Comparison to SlowFast}
SlowFast~\cite{feichtenhofer_iccv_2019} proposes a two-stream CNN architecture that takes frames sampled at two different frame rates. The ``Slow’’ pathway, built with a larger encoder, processes the low frame rate stream to capture the semantics of the scene while the ``Fast’’ pathway that takes in high frame rate inputs is used to capture motion information. 
To make a fair comparison, we implement~\cite{feichtenhofer_iccv_2019} in the context of transformers where we use ``Base'' and ``Tiny'' models as the encoders for the Slow and Fast paths respectively and use CVA for lateral connections. The Slow path takes four frames as inputs sampled with a temporal stride of 16 and the Fast path takes 16 frames sampled with a stride of 4.
As SlowFast captures multiscale temporal information by varying the frame rate to the two streams, the temporal duration for the tubelets is set to 1 in this case.
Table~\ref{tab:slowfast} shows that our method is significantly more accurate than the SlowFast method whilst also using fewer FLOPs.

\subsection{Comparison to the state of the art}
\label{sec:sota}
\input{sota}

We compare to the state-of-the-art across six different datasets.
We evaluate models with four temporal- and three spatial-views per video clip, following~\cite{arnab2021vivit}. 
To make the notation more concise, we now use MTV-B to refer to B/2+S/4+Ti/8, MTV-L to refer to L/2+B/4+S/8+Ti/16 and MTV-H to refer to H/2+B/4+S/8+Ti/16.
Except for Kinetics, all our models start from a Kinetics 400 checkpoint and then are fine-tuned on the target datasets following~\cite{arnab2021vivit,patrick2021keeping,fan2021multiscale}.

\paragraph{Accuracy/computation trade-offs}
Figure~\ref{fig:comparison_to_vivit} compares our proposed MTV to its single-view counterpart, ViViT Factorized Encoder (FE)~\cite{arnab2021vivit} at every model scale on Kinetics 400.
We compare to ViViT-FE using tubelets with a temporal dimension of $t = 2$, as the authors obtained the best performance with this.

We can control the complexity of MTV by increasing or decreasing $t$ used in each view.
For example, increasing $t$ from 2 to 4 for the smallest view (and proportionally increasing $t$ for all other views) will roughly reduce the input tokens by half for each view, and thus halve the total FLOPs for processing each input.
Our method with $t=4$ for the smallest view consistently achieves higher accuracy than ViViT-FE at every complexity level while using fewer FLOPs, indicated by the green arrows pointing to the upper-left in Fig.~\ref{fig:flops}.
This further validates that processing more views in parallel enables us to achieve larger accuracy improvements than increasing the number of input tokens.
If we set $t = 2$ as in ViViT-FE, we use additional FLOPs, but increase significantly in accuracy too, as indicated by the green arrow pointing to the upper-right in Fig.~\ref{fig:flops}.

Furthermore, note how our B/2 model (transformer depth of 12 layers) outperforms ViViT-L/2 (24 layers), whilst using less FLOPs.
Similarly, our L/2 model outperforms ViViT-H/2.
This shows that we can achieve greater accuracy improvements by processing multiple views in parallel than by increasing the depth for processing a single view.

Finally, note that Fig.~\ref{fig:latency} shows that our conclusions are also consistent when using the inference time to measure our model's efficiency.
Appendix~\ref{sec: comparison_to_vivit} also shows that these trends also hold when using an unfactorized~\cite{arnab2021vivit} backbone architecture of ViViT and MTV.

\paragraph{Kinetics}
We compare to methods that are pretrained on ImageNet-1K, ImageNet-21K~\cite{deng2009imagenet} and those that do not utilize pretraining at all in the first part of Tab.~\ref{tab:sota}.
In the second part of the tables, we compare to methods that are pretrained on web-scale datasets such as Instagram 65M~\cite{ghadiyaram2019large}, JFT-300M~\cite{300m_iccv17}, JFT-3B~\cite{zhai2021scaling}, WTS~\cite{stroud2020learning}, Florence~\cite{yuan2021florence} or HowTo100M~\cite{miech2019howto100m}.
Observe that we achieve state-of-the-art results both with and without web-scale pretraining.

On Kinetics 400, our ImageNet-21K pretrained ``Base'' model improves the ``Large'' ViViT-FE model~\cite{arnab2021vivit}, which corresponds to a  deeper, single-view equivalent of our model by 0.1\% and 1.2\% in Top-1 and Top-5 accuracy, whilst using 40\% of the total FLOPs.
Our higher resolution version improves further by 0.7\% and 1.4\% while still using slightly fewer FLOPs.
On Kinetics 600, our ``Base'' model scores second to~\cite{kondratyuk2021movinets} whose model structure is derived using architecture search on Kinetics 600 itself.
We show significant improvements over~\cite{kondratyuk2021movinets} on both Kinetics 400 and 700 for which the architecture of~\cite{kondratyuk2021movinets} was not directly optimized for.

When using additional JFT-300M pretraining, our ``Huge'' model outperforms other recent transformer models using the same pretraining dataset~\cite{arnab2021vivit, ryoo2021tokenlearner}.
And when we utilize the Weak Textual Supervision (WTS) dataset of~\cite{stroud2020learning} for pre-training, we substantially advance the best reported results on Kinetics:
On Kinetics 400, we achieve a Top-1 accuracy of 89.9\%, which improves upon the previous highest result (CoVeR~\cite{zhang2021co}) by 2.7\%.
Similarly, on Kinetics 600, we achieve a Top-1 of 90.3\%, which is an absolute improvement of 2.4\% on~\cite{zhang2021co}.
On Kinetics 700, we achieve 83.4\%, which improves even further by 3.6\% over~\cite{zhang2021co}.
We also improve upon R3D-RS~\cite{du2021revisiting}, which also used WTS pretraining, by 6.4\% and 6.0\% on Kinetics-400 and -600. 

\paragraph{Epic-Kitchens-100}
Following the standard protocol \cite{damen2021rescaling}, we report Top-1 action-, verb- and noun-accuracies with action accuracy being the primary metric.
Our results are averaged over $4\times1$ crops as additional spatial crops did not help.
Both our MTV-B and MTV-B(320p) significantly improve the previous state-of-the-art on noun classes, and MTV-B(320p) achieves a new state-of-the-art of 48.6\% on actions. With WTS pretraining and increasing resolution, we improved the results to 50.5\%. 
We found that additional data augmentation (detailed in Appendix \ref{sec: hparams}) has to be used to achieve good performance (as also observed by~\cite{arnab2021vivit, patrick2021keeping}) as this is the smallest dataset of all six with 67,000 training examples.

\paragraph{Something-Something V2}
This dataset consists of class labels such as ``move to left'' and ``pointing to right''~\cite{goyal_iccv_2017}.
As the model needs to explicitly reason about direction, we do not perform random horizontal or vertical flipping as data augmentation on this dataset as also done by~\cite{fan2021multiscale}.
We improve substantially over ViViT-L-FE~\cite{arnab2021vivit}, which corresponds to a deeper single-view equivalent of our model 
by 2.6\%, and also improve upon MFormer~\cite{patrick2021keeping} by 0.4\%.

\paragraph{Moments in Time} 
Our MTV-L model significantly improves over the previous state-of-the-art~\cite{kondratyuk2021movinets} by 1.5\% in Top-1 accuracy.
Moreover, our model with ImageNet-21K pretraining even outperforms VATT~\cite{akbari2021vatt}, which was pretrained on HowTo100M~\cite{miech2019howto100m}, a dataset consisting of around 100M video clips.
When using WTS pre-training, we improve our accuracy even further, achieving 47.2\%.

%% file: ablation.tex
\begin{table*}[t]
    \vspace{-0.5\baselineskip}
	\begin{subtable}[t]{.31\linewidth}
		\centering
		\caption{Effects of different model-view assignments.
		}
		\vspace{-0.1\baselineskip}
    	\setlength{\tabcolsep}{4pt} %
		\renewcommand*{\arraystretch}{1.11}  %
		\scriptsize{
			\begin{tabular}{  l c c c }
			    \toprule
                Model variants & GFLOPs & MParams & Top-1 \\
                \midrule
                B/8+Ti/2 & 81 & 161 &77.3  \\ %
                B/2+Ti/8 & 337 & 221 & 81.3 \\ %
                \midrule
                B/8+S/4+Ti/2 & 202 & 250 & 78.5 \\ 
                B/2+S/4+Ti/8 & 384 & 310 & 81.8 \\ %
                B/4+S/8+Ti/16 & 195 & 314 & 81.1  \\ %
                \bottomrule
             \end{tabular}
		}
		\label{tab:model_view_assignment}
		\vspace{1.\baselineskip}
		\centering
		\caption{Effects of the same model applied to different views.}
		\vspace{-0.3\baselineskip}
		\scriptsize{
			\begin{tabular}{  l c c c }
			    \toprule
                Model variants & GFLOPs & MParams & Top-1 \\
                \midrule
                B/4+S/8+Ti/16 & 195 & 314 & 81.1 \\ %
                B/4+B/8+B/16 & 324 & 759 & 81.1\\ %
                \midrule
                B/2+Ti/8 & 337 & 221 & 81.3 \\ %
                B/2+B/8 & 448 & 465 & 81.5\\ %
                \midrule
                B/2+S/4+Ti/8 & 384 & 310 & 81.8 \\ %
                B/2+B/4+B/8 & 637 & 751 & 81.7 \\ %
                \bottomrule
             \end{tabular}
		}
		\label{tab:same_model_different_views}
	\end{subtable}
  	\hfill
  	\begin{subtable}[t]{.37\linewidth}
		\centering
  		\caption{Comparison of different cross-view fusion methods.}
  		\vspace{-0.1\baselineskip}
  		\setlength{\tabcolsep}{4pt} %
  		\scriptsize{
	  		\begin{tabular}{  l  c  c  c  l }
	  		    \toprule
                Model variants & Method & GFLOPs & MParams & Top-1 \\
                \midrule
                B/4 & \multirow{3}{*}{N/A} & 145 & 173 & 78.3 \\ %
                S/8 &  & 20 & 60 & 74.1 \\ %
                Ti/16 &  & 3 & 13 & 67.6 \\ %
                \midrule
                \multirow{5}{*}{B/4+S/8+Ti/16} & Ensemble & 168 & 246 & 77.7   \\ %
                & Late fusion & 187 & 306 & 80.6  \\ %
                & MLP & 202 & 323 & 80.6  \\ %
                & Bottleneck & 188 & 306 & 81.0  \\ 
                & CVA & 195 & 314 & \textbf{81.1}  \\ %
                \bottomrule
    \end{tabular}
	\label{tab:fusion_methods}
  	}
    \vspace{1.05\baselineskip}
    \centering
	\scriptsize{
	\caption{Comparison to SlowFast multi-resolution method. %
	}
	\vspace{-0.1\baselineskip}
    \begin{tabular}{  l  c  c  c }
      \toprule
      Model variants & GFLOPs & MParams & Top-1 \\
      \midrule
      \multicolumn{3}{l}{\textit{SlowFast (transformer backbone)}} \\
      Slow-only (B) & 79 & 87 & 78.0 \\ %
      Fast-only (Ti) & 63 & 6 & 74.6 \\ %
      Slowfast (B+Ti) & 202 & 105 & 79.7 \\ \midrule %
      B/4+Ti/16 (ours) & 168 & 224  & \textbf{80.8} \\ %
      \bottomrule
    \end{tabular}}
    \label{tab:slowfast}
  	\end{subtable}
  	\hfill
  	\begin{subtable}[t]{.3\linewidth}
  		\setlength{\tabcolsep}{6pt} %
  		\centering
  		\caption{Effects of increasing number of views.}
  		\vspace{-0.1\baselineskip}
  		\scriptsize{
  			\begin{tabular}{  l  c  l }
  			    \toprule
                Model variants & GFLOPs & Top-1 \\
                \midrule
                B/4 & 145 & 78.3  \\ %
                B/4+Ti/16 & 168 & 80.8 (\textcolor{green}{+2.5}) \\ %
                B/4+S/8+Ti/16 & 195 & 81.1 (\textcolor{green}{+2.8}) \\ %
                \midrule
                B/4 (14) & 168 &  78.1 (\textcolor{red}{-0.2}) \\ %
                B/4 (17) & 203 & 78.4 (\textcolor{green}{+0.1}) \\ %
                \bottomrule
               \end{tabular}
  			\label{tab:num_views}
  		}
  		\vspace{1.4\baselineskip}
  		\setlength{\tabcolsep}{4pt} %
  		\centering
	    \scriptsize{
	    \caption{Effects of applying CVA at different layers.} %
	    \vspace{-0.1\baselineskip}
        \begin{tabular}{  c  c  c  c }
        \toprule
        Fusion layers & GFLOPs & MParams & Top-1 \\
        \midrule
        0 & \multirow{3}{*}{195} & \multirow{3}{*}{314} &  80.96 \\
        5 & &  & 81.08 \\
        11 & &  & 81.00 \\ \midrule
        0, 1 & \multirow{4}{*}{203} & \multirow{4}{*}{323} & 80.91 \\
        5, 6 &  &  & 80.96 \\
        10, 11 &  &  & 80.81 \\
        5, 11 &  & & \textbf{81.14} \\
        \midrule
        0, 5, 11 & 210 & 331 & 80.95 \\
        \bottomrule
       \end{tabular}
       \label{tab:cva_layers}
       }
	\end{subtable}
	\vspace{-0.5\baselineskip}
	\caption{Ablation studies of our method.
	(a) Assigning larger models to smaller tubelet sizes achieves the highest accuracy.
	(b) We apply the same ``Base'' encoder to all views, and show that there is minimal accuracy difference to the alternatives from (a), but a large increase in computation.
	(c) A comparison of different cross-view fusion methods, shows that Cross-View Attention (CVA) is the best.
	The ``Ensemble'' and ``late fusion'' baselines are detailed in the text.
	(d) We compare our approach to the alternate temporal multi-resolution method of \cite{feichtenhofer_iccv_2019}, implemented in the context of transformers, and show signficant improvements.
	(e) We achieve substantial accuracy by adding more views, and this improvement is larger than that obtained by adding more layers to a single encoder.
	(f) The optimal fusion layers are at the middle and late stages of the network. %
		}
	\label{tab:ablation}
	\vspace{-\baselineskip}
\end{table*}

%% file: vivit_vs_mtv.tex
\begin{figure*}
     \centering
     \begin{subfigure}[b]{0.47\textwidth}
         \centering
         \includegraphics[width=\textwidth]{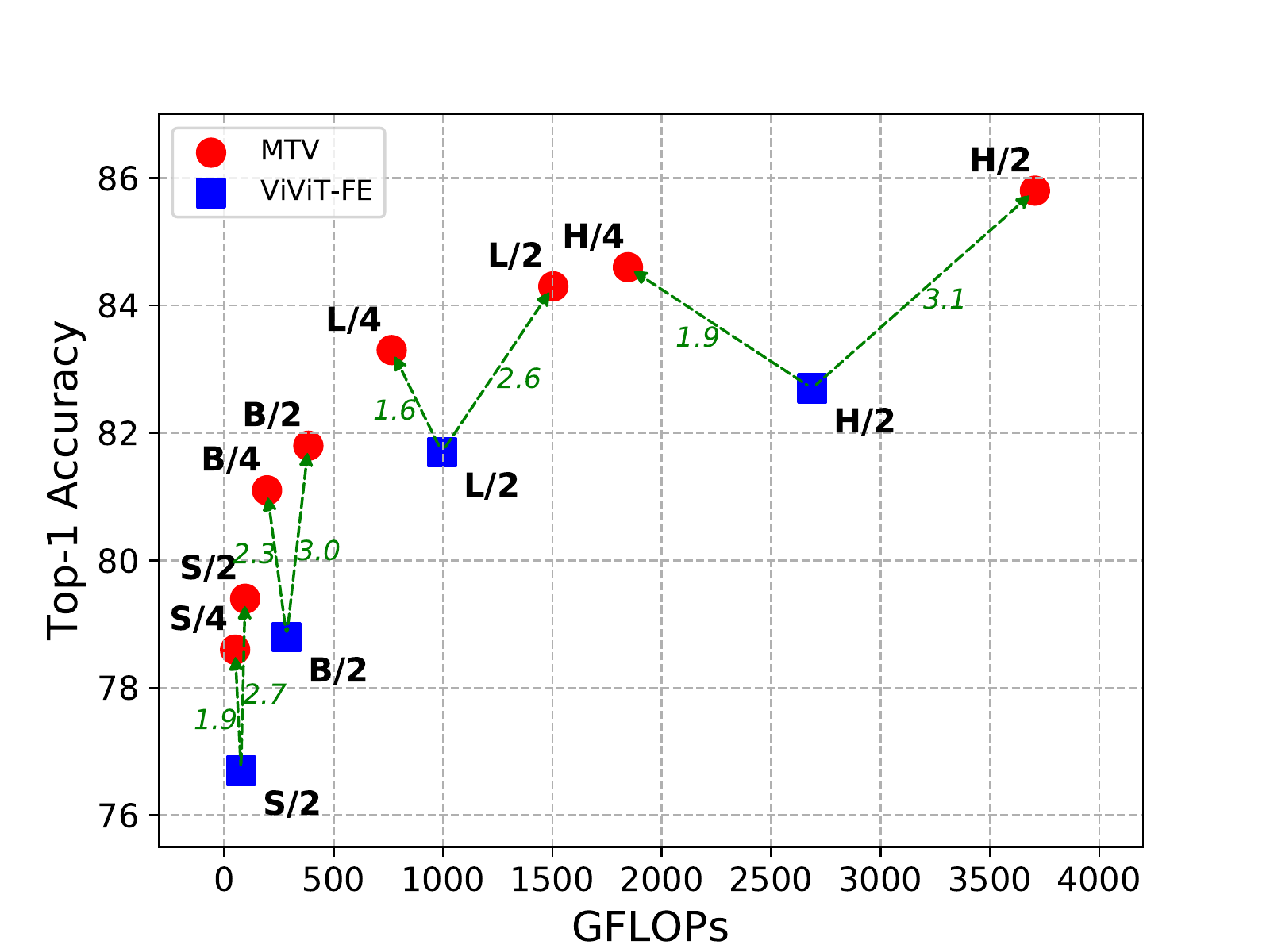}
         \caption{Accuracy[\%] - GFLOPs comparison between MTV and ViViT-FE.}
         \label{fig:flops}
     \end{subfigure}
     \hfill
     \begin{subfigure}[b]{0.47\textwidth}
         \centering
         \includegraphics[width=\textwidth]{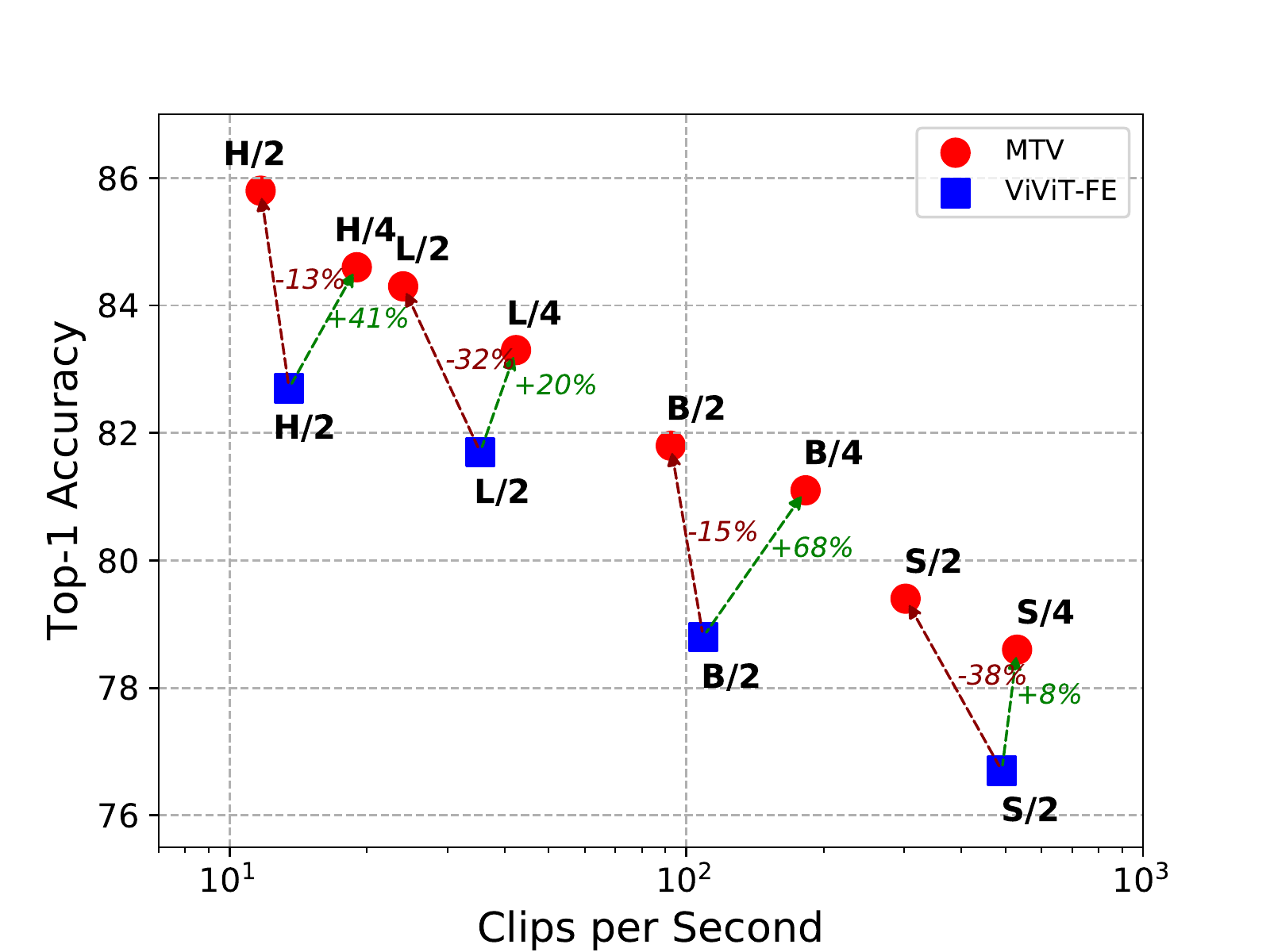}
        \caption{Accuracy[\%] - Throughput comparison between MTV and ViViT-FE.}
         \label{fig:latency}
     \end{subfigure}
        \caption{
        Accuracy/computation trade-off between ViViT-FE~\cite{arnab2021vivit} (blue) and our MTV (red). Figure~\ref{fig:flops} shows that MTV is consistently better and requires less FLOPs than ViViT-FE to achieve higher accuracy across different model scales (shown by the dotted green arrows pointing upper-left). With additional FLOPs, MTV shows larger accuracy gains (shown by the dotted green arrows pointing upper-right).
        Similarly, Fig.~\ref{fig:latency} shows that MTV can have higher throughput than ViVIT-FE, whilst still improving its accuracy, across all model scales.
        All speed comparisons are measured with the same hardware (Cloud TPU-v4), whilst the accuracy is computed from $4\times 3$ view testing.
        }
        \label{fig:accuracy/complexity tradeoff}
    \label{fig:comparison_to_vivit}
    \vspace{-\baselineskip}
\end{figure*}

%% file: sota.tex
\begin{table*}[t]
    \vspace{-0.5\baselineskip}
	\caption{Comparisons to state-of-the-art. For ``views'', $x \times y$ denotes $x$ temporal views and $y$ spatial views. We report the total TFLOPs 
	to process all spatio-temporal views.
	We use shorter notation, MTV-B, L, H to denote variants, B/2+S/4+Ti/8, L/2+B/4+S/8+Ti/16, and H/2+B/4+S/8+Ti/16, respectively. Models use a spatial resolution of $224\times224$, unless explicitly stated by MTV ($x$p), which refers to a spatial resolution of $x\times x$. 
	Models are pretrained on ImageNet-21K unless explicitly stated in parenthesis.
	}
	\vspace{-0.5\baselineskip}
	\begin{subtable}[t]{.39\linewidth}
		\centering
		\caption{Kinetics 400}
    	\setlength{\tabcolsep}{3pt} %
		\renewcommand*{\arraystretch}{1.10}  %
		
		\vspace{-0.3\baselineskip}
		\scriptsize{
			\begin{tabular}{lcccc}
				\toprule
				Method 																			 & Top 1                & Top 5         & Views 	& TFLOPs \\
				\midrule
				TEA~\cite{li_tea_cvpr_2020}												& 76.1					& 92.5 & $10 \times 3$ & 2.10 \\  %
				TSM-ResNeXt-101~\cite{lin_tsm_cvpr_2019}						  & 76.3				 & -- &  -- & -- \\%
				I3D NL~\cite{wang2018nonlocal}										 & 77.7                  & 93.3         		 &  $10 \times 3$ & 10.77     \\ 
				VidTR-L~\cite{zhang2021vidtr} & 79.1 & 93.9 & $10 \times 3$ & 10.53 \\
				LGD-3D R101~\cite{qiu2019learning}							&  79.4				    & 94.4			 		&  --	& --					\\  
				SlowFast R101-NL~\cite{feichtenhofer_iccv_2019}       		&  79.8                 &  93.9                   & $10 \times 3$    & 7.02   \\  %
				X3D-XXL~\cite{feichtenhofer_cvpr_2020}      					&  80.4					&  94.6			  		& $10 \times 3$   & 5.82   \\
				OmniSource~\cite{duan2020omni} & 80.5 & 94.4 & -- & -- \\
				TimeSformer-L~\cite{bertasius_arxiv_2021} & 80.7				& 94.7					& $1 \times 3$ & 7.14 \\
				MFormer-HR~\cite{patrick2021keeping} & 81.1 & 95.2 & $10 \times 3$ & 28.76 \\
				MViT-B~\cite{fan2021multiscale}					  	& 81.2				& 95.1					& $3 \times 3$ & 4.10 \\
				MoViNet-A6~\cite{kondratyuk2021movinets} & 81.5 	&  \textbf{95.3}  & $1 \times 1$ & 0.39  \\  %
				ViViT-L FE~\cite{arnab2021vivit} & 81.7 	&  93.8  & $1 \times 3$ &  11.94 \\  %
				\textbf{MTV-B} & \textbf{81.8}  & 95.0 	& $4 \times 3$ & 4.79 \\ %
				\textbf{MTV-B} (320p) & \textbf{82.4}  & 95.2 	& $4 \times 3$ & 11.16 \\ %
				\midrule
				\multicolumn{4}{l}{\textit{Methods with web-scale pretraining}}                                \\ 
				VATT-L~\cite{akbari2021vatt} (HowTo100M)  &  82.1 &  95.5 	& $4 \times 3$ & 29.80 \\  %
				ip-CSN-152~\cite{tran_iccv_2019} (IG) 			  &  82.5					& 95.3			&  $10 \times 3$ & 3.27	  \\
				R3D-RS (WTS)~\cite{du2021revisiting} & 83.5 & -- & $10 \times 3$ & 9.21  \\
				OmniSource~\cite{duan2020omni} (IG) & 83.6 & 96.0 & -- & -- \\
				ViViT-H~\cite{arnab2021vivit} (JFT) 														&  84.9 &  95.8 	& $4 \times 3$ & 47.77 \\  %
				TokenLearner-L/10~\cite{ryoo2021tokenlearner} (JFT) 														&  85.4 &  96.3 	& $4 \times 3$ & 48.91 \\  %
				Florence~\cite{yuan2021florence} (FLD-900M) & 86.5 & 97.3 & $4 \times 3$ & -- \\
				CoVeR (JFT-3B)~\cite{zhang2021co} & 87.2 & -- & $1 \times 3$ & -- \\
				\textbf{MTV-L} (JFT) 														&  84.3 &  96.3 	& $4 \times 3$ & 18.05 \\  %
				\textbf{MTV-H} (JFT) 														&  85.8 &  96.6 	& $4 \times 3$ & 44.47 \\ %
				\textbf{MTV-H} (WTS) 														&  \textbf{89.1} &  \textbf{98.2} 	& $4 \times 3$ & 44.47 \\ %
				\textbf{MTV-H} (WTS 280p) 														&  \textbf{89.9} &  \textbf{98.3} 	& $4 \times 3$ & 73.57 \\ %
				\bottomrule
			\end{tabular}
		}
		\label{tab:sota_kinetics400}
	\end{subtable}
  	\hfill
  	\begin{subtable}[t]{.29\linewidth}
		\centering
  		\caption{Kinetics 600}
  		\setlength{\tabcolsep}{4pt} %
		\vspace{-0.3\baselineskip}
  		\scriptsize{
	  		\begin{tabular}{lcc}
	  			\toprule
	  			Method 																			 & Top 1                & Top 5      \\ %
	  			\midrule
	  			SlowFast R101-NL~\cite{feichtenhofer_iccv_2019}       		&  81.8                     &  95.1   \\ %
	  			X3D-XL~\cite{feichtenhofer_cvpr_2020}      						 &  81.9					&  95.5		\\ %
	  			TimeSformer-L~\cite{bertasius_arxiv_2021}				  & 82.2				& 95.6		\\ %
	  			MFormer-HR~\cite{patrick2021keeping} & 82.7 & 96.1  \\
	  			ViViT-L FE~\cite{arnab2021vivit} & 82.9 	&  94.6 \\  %
	  			MViT-B~\cite{fan2021multiscale}					  & 83.8				& 96.3		\\ %
	  			MoViNet-A6~\cite{kondratyuk2021movinets} & \textbf{84.8} 	&  \textbf{96.5} \\
	  			\textbf{MTV-B} & 83.6  & 96.1 \\ %
				\textbf{MTV-B} (320p) & 84.0  & 96.2 \\ %
	  			\midrule
  				R3D-RS (WTS)~\cite{du2021revisiting} & 84.3 & --  \\
	  			ViViT-H~\cite{arnab2021vivit} (JFT) & 85.8 & 96.5 \\ %
	  			TokenLearner-L/10~\cite{ryoo2021tokenlearner} (JFT) &  86.3 &  97.0 \\  %
	  			Florence~\cite{yuan2021florence} (FLD-900M) & 87.8 & 97.8 \\
	  			CoVeR (JFT-3B)~\cite{zhang2021co} & 87.9 & -- \\
	  			\textbf{MTV-L} (JFT) &  85.4 &  96.7 \\ %
	  			\textbf{MTV-H} (JFT) &  86.5 &  97.3 \\ %
	  			\textbf{MTV-H} (WTS) &  \textbf{89.6} &  \textbf{98.3} \\ %
	  			\textbf{MTV-H} (WTS 280p) &  \textbf{90.3} &  \textbf{98.5} \\ %
	  			\bottomrule
	  		\end{tabular}
	  		\label{tab:sota_kinetics600}
  		}
  		\vspace{0.6\baselineskip} %
		\centering
		\caption{Something-Something v2}
		\vspace{-0.1\baselineskip}
		\setlength{\tabcolsep}{6pt} %
		\scriptsize{
			\begin{tabular}{lcc}
				\toprule
				Method 												  & Top 1                & Top 5   \\
				\midrule
				SlowFast R50~\cite{feichtenhofer_iccv_2019,wu_multigrid_cvpr_2020}		& 61.7 & --    \\
				TimeSformer-HR~\cite{bertasius_arxiv_2021}					& 62.5 & -- \\
				VidTR~\cite{zhang2021vidtr} & 63.0 & -- \\
				ViViT-L FE~\cite{arnab2021vivit}		& 65.9  & 89.9 \\ %
				MViT~\cite{fan2021multiscale} & 67.7 & 90.9 \\
				MFormer-L~\cite{patrick2021keeping} & 68.1 & \textbf{91.2}  \\
				\textbf{MTV-B} & 67.6 & 90.1 \\ %
				\textbf{MTV-B} (320p) & \textbf{68.5} & 90.4 \\ %
				\bottomrule
			\end{tabular}
			\label{tab:sota_ssv2}
		}
  	\end{subtable}
  	\hfill
  	\begin{subtable}[t]{.28\linewidth}
  		\setlength{\tabcolsep}{4pt} %
  		\centering
  		\caption{Kinetics 700}
  		\vspace{-0.3\baselineskip}
  		\setlength{\tabcolsep}{6pt} %
  		\scriptsize{
  			\begin{tabular}{lcc}
  				\toprule
  				& Top 1 & Top 5 \\ 
  				\midrule
  				VidTR-L~\cite{zhang2021vidtr} & 70.2 & -- \\
  				SlowFast R101~\cite{feichtenhofer_iccv_2019} 	&  71.0 & 89.6       \\
  				MoViNet-A6~\cite{kondratyuk2021movinets} 	&  72.3     &  --     \\
  				\textbf{MTV-L} & \textbf{75.2} & \textbf{91.7} \\  %
  				\midrule
	  			CoVeR (JFT-3B)~\cite{zhang2021co} & 79.8 & -- \\
  				\textbf{MTV-H} (JFT) & 78.0 & 93.3 \\  %
  				\textbf{MTV-H} (WTS) & \textbf{82.2} & \textbf{95.7} \\  %
  				\textbf{MTV-H} (WTS 280p) & \textbf{83.4} & \textbf{96.2} \\  %
  				\bottomrule
  			\end{tabular}
  			\label{tab:sota_kinetics700}
  		}
  		\vspace{0.18\baselineskip} %
  		\centering
  		\caption{Epic-Kitchens-100 Top 1 accuracy}
  		\vspace{-0.2\baselineskip}
  		\setlength{\tabcolsep}{4pt} %
		\scriptsize{
			\begin{tabular}{lccc}
				\toprule
				Method 													 & Action & Verb  & Noun  \\
				\midrule
				ViViT-L FE~\cite{arnab2021vivit} & 44.0 & 66.4 & 56.8 \\ %
				MFormer-HR~\cite{patrick2021keeping} & 44.5 & 67.0 & 58.5  \\
				MoViNet-A6~\cite{kondratyuk2021movinets} & 47.7 & \textbf{72.2} & 57.3 \\ %
				\textbf{MTV-B} & 46.7 & 67.8 & \textbf{60.5} \\ %
			    \textbf{MTV-B} (320p) & \textbf{48.6} & 68.0 & \textbf{63.1} \\ %
			    \midrule
			    \textbf{MTV-B} (WTS 280p) & \textbf{50.5} & 69.9 & \textbf{63.9} \\ 
				\bottomrule
			\end{tabular}
		}
		\label{tab:sota_epic_kitchens}
  		\centering
  		\caption{Moments in Time}
  		\vspace{-0.1\baselineskip}
  		\setlength{\tabcolsep}{6pt} %
  		\scriptsize{
  			\begin{tabular}{lcc}
  				\toprule
  				& Top 1 & Top 5 \\ 
  				\midrule
  				AssembleNet-101~\cite{ryoo2019assemblenet} 	&  34.3     &  62.7     \\
  				ViViT-L FE~\cite{arnab2021vivit} 	& 38.5 & 64.1 \\  %
  				MoViNet-A6~\cite{kondratyuk2021movinets} 	&  40.2     &  --     \\
  				\textbf{MTV-L}	& \textbf{41.7} & \textbf{69.7} \\  %
  				\midrule
  				VATT-L (HT100M)~\cite{akbari2021vatt}	&  41.1     &  67.7     \\
  				\textbf{MTV-H} (JFT)	& \textbf{44.0} & \textbf{70.2} \\  %
  				\textbf{MTV-H} (WTS)	& \textbf{45.6} & \textbf{74.7} \\  %
  				\textbf{MTV-H} (WTS 280p)	&
  				\textbf{47.2} & \textbf{75.7} \\  
  				\bottomrule
  			\end{tabular}
  			\label{tab:sota_moments_in_time}
  		}

	\end{subtable}
	\label{tab:sota}
	\vspace{-\baselineskip}
\end{table*}

%% file: conc.tex
\section{Conclusion}
We have presented a simple method for capturing multi-resolution temporal context in transformer architectures, based on processing multiple ``views'' of the input video in parallel.
We have demonstrated that our approach performs better, in terms of accuracy/computation trade-offs than increasing the depth of current single-view architectures. 
Furthermore, we have achieved state-of-the-art results on six popular video classification datasets.
These results were then further improved with large-scale pretraining~\cite{300m_iccv17, stroud2020learning}.

\paragraph{Limitations and future work} Although we have improved upon the state-of-the-art, there is still a large room for improvement on datasets other than Kinetics.
Furthermore, we have relied on models pretrained on large image- or video-datasets for initialization.
Reducing this dependence on supervised pretraining is a clear avenue of future research.
We have conducted thorough ablations on standard transformer architectures~\cite{dosovitskiy2020image, arnab2021vivit}, and will investigate if our approach is complementary to recent, spatial-pyramid based multiscale transformer encoders such as MViT~\cite{fan2021multiscale} and Swin~\cite{liu2021swin}.

\paragraph{Societal impact} Video classification models can be used in a wide range of applications.
We are unaware of all potential applications, but are mindful that each application has its own merits, and that also depends on the intentions of the individuals building and using these systems.
We also note that training datasets may contain biases that models trained on them are unsuitable for certain applications.

%% file: supplementary.tex
\twocolumn[
\centering
\Large
\vspace{-1.5em}
] %
\appendix

\section{Additional experiments}
In this Appendix, we provide additional experimental details.
Section~\ref{sec: comparison_to_vivit} provides accuracy-FLOPs and accuracy-throughput comparison between two model variants of ViViT and MTV.
Section~\ref{sec: spatial} provides the effect of spatial resolution of tubelets.
Section~\ref{sec: hparams} and Section~\ref{sec: model_configs} provides details of our training hyperparameters and model configurations used in our experiments.

\subsection{Changing transformer encoder architecture}
\label{sec: comparison_to_vivit}

We present additional results by changing the transformer architecture used within our multiview encoder.
Specifically, we use the unfactorized ViViT transformer encoder (Model 1 of~\cite{arnab2021vivit}).
In this variant, each transformer encoder layer computes self-attention over all spatio-temporal tokens.
This makes our multiview transformer encoder cover a wide range of spatial and temporal dimensions across different views. A one-layer MLP with hidden dimension of $3072$ is used as the global encoder for our unfactorized MTV model.

\begin{figure*}
     \vspace{-0.5\baselineskip}
     \centering
     \begin{subfigure}[b]{0.49\textwidth}
         \centering
         \includegraphics[width=\textwidth]{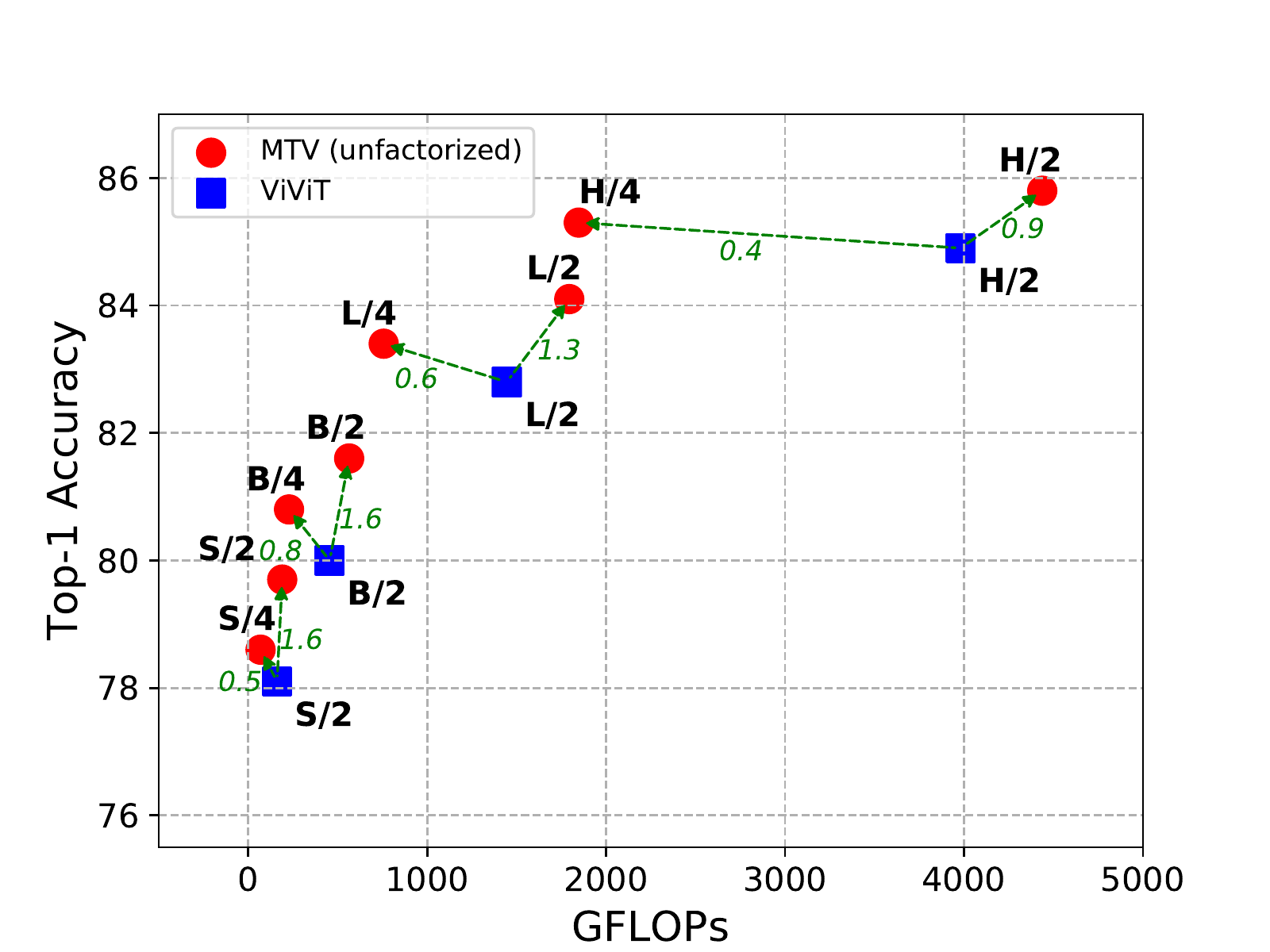}
         \caption{Accuracy[\%] - GFLOPs comparison between MTV (unfactorized) and ViViT.}
         \label{fig:mtv_vs_vivit_3d_flops}
     \end{subfigure}
     \hfill
     \begin{subfigure}[b]{0.49\textwidth}
         \centering
         \includegraphics[width=\textwidth]{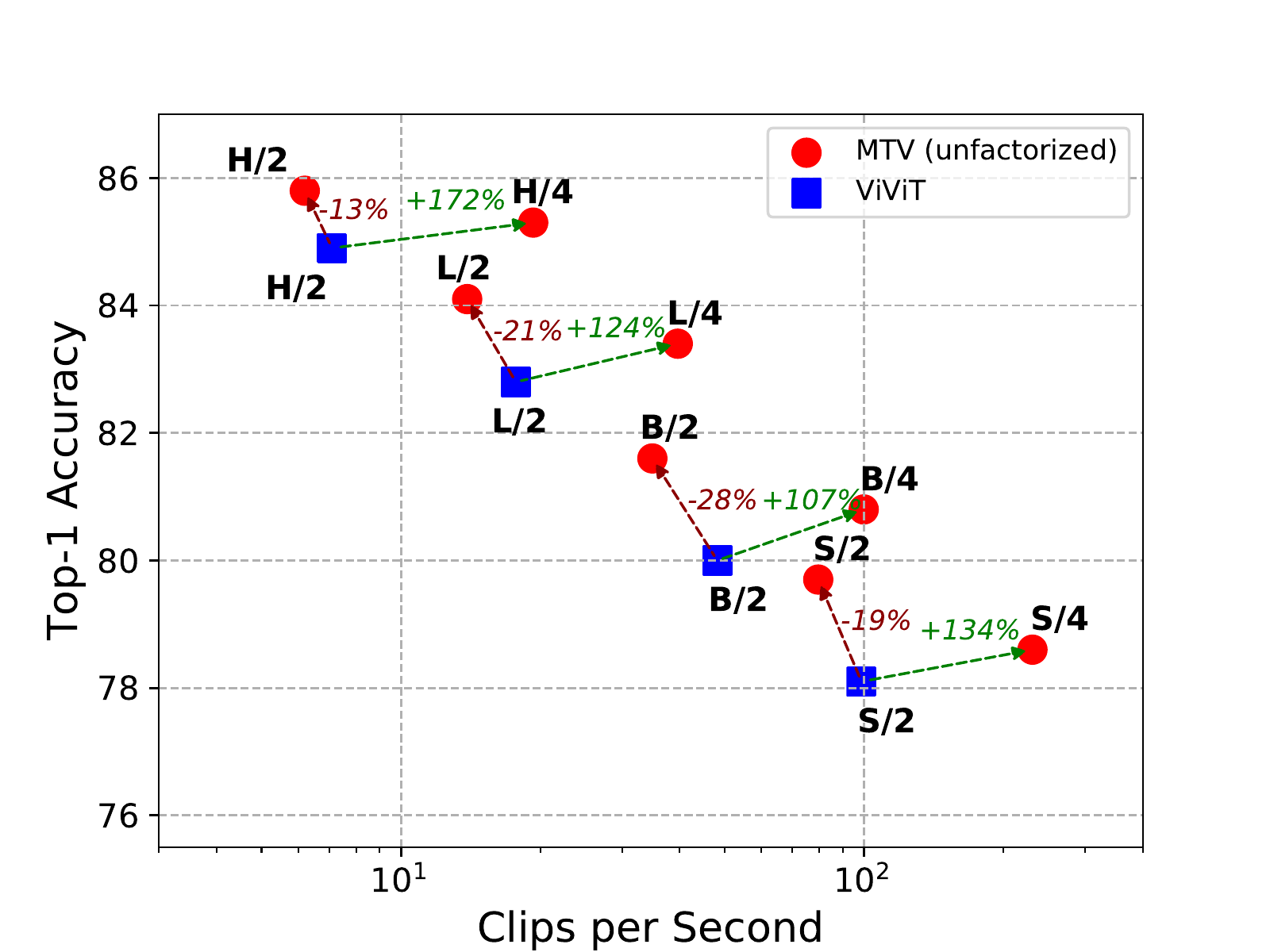} %
         \caption{Accuracy[\%] - Throughput comparison between MTV (unfactorized) and ViViT.}
         \label{fig:mtv_vs_vivit_3d_latency}
     \end{subfigure}
          \begin{subfigure}[b]{0.49\textwidth}
         \centering
         \includegraphics[width=\textwidth]{mtv_fe_vs_vivit_fe.pdf}
         \caption{Accuracy[\%] - GFLOPs comparison between MTV and ViViT-FE.}
         \label{fig:mtv_vs_vivit_fe_flops}
     \end{subfigure}
     \hfill
     \begin{subfigure}[b]{0.49\textwidth}
         \centering
         \includegraphics[width=\textwidth]{speed_mtv_fe_vs_vivit_fe.pdf}
         \caption{Accuracy[\%] - Throughput comparison between MTV and ViViT-FE.}
         \label{fig:mtv_vs_vivit_fe_latency}
     \end{subfigure}
        \caption{Accuracy/complexity trade-off between ViViT / ViViT-FE~\cite{arnab2021vivit} (blue) and our MTV (unfactorized) / MTV (red). MTV (unfactorized) is consistently better and requires less FLOPs (see Fig. \ref{fig:mtv_vs_vivit_3d_flops}) than ViViT to achieve higher accuracy across different model scales (indicated by the dotted green arrows pointing upper-left). With additional FLOPs, MTV shows larger accuracy gains (shown by the dotted green arrows pointing upper-right). The lower number of FLOPs is translated to higher throughput (clips per second), as indicated by the green arrows in Fig. \ref{fig:mtv_vs_vivit_3d_latency}.
        Note how MTV can more than double the throughput of ViViT unfactorized, whilst still improving its accuracy, across all model scales. %
        Similar findings are also observed by the comparison between ViViT-FE and MTV model in Fig. \ref{fig:mtv_vs_vivit_fe_flops} and Fig. \ref{fig:mtv_vs_vivit_fe_latency}.
        Note that Fig.~\ref{fig:mtv_vs_vivit_fe_flops} appeared as Figure 3 in the main paper, and is included here for clarity and consistency.
        All speed comparisons are measured with the same hardware (Cloud TPU-v4). %
        The complexity is for a single $32 \times 224 \times 224 \times 3$ input video (denoted as $T \times H  \times W \times C$), and the accuracy is obtained by $4\times 3$ view testing. 
        }
    \label{fig: mtv_vs_vivit}
    \vspace{-\baselineskip}
\end{figure*}

As shown in Fig.~\ref{fig: mtv_vs_vivit}, MTV (unfactorized) consistently outperforms its single-view counterpart (\emph{i.e.} ViViT unfactorized) for every scale (see Fig.~\ref{fig:mtv_vs_vivit_3d_flops}) and corresponds to a better accuracy-throughput curve as shown in Fig.~\ref{fig:mtv_vs_vivit_3d_latency}. 
Note how MTV can more than double the throughput of ViViT unfactorized, whilst still improving its accuracy, for each model scale.
Specifically, MTV (unfactorized) H/4+B/8+S/16+Ti/32 model leads to a significant speed-up by $172\%$ while still keeping a higher accuracy of $0.4\%$ improvement compared to ViViT-H. 

Moreover, we report the accuracy-throughput comparison between MTV and ViViT factorized model (ViViT-FE) in Fig.~\ref{fig:mtv_vs_vivit_fe_latency}. Note that the accuracy-FLOPs comparison is already reported in paper Section~4.3. 
The improvements in accuracy-throughput, and accuracy-FLOPs remain significant in this setting.

Note that the unfactorized ViViT transformer encoder, which attends to all spatio-temporal tokens, is less efficient than the Factorized Encoder architecture that we used in the main paper.
However, we achieve larger relative improvements in accuracy/computation trade-offs compared to the corrsponding single-view ViViT baseline when using this encoder architecture.

\subsection{Spatial resolution of tubelets} \label{sec: spatial}
We study the effect of the spatial resolution of tubelets in Tab.~\ref{tab:spatial_ablations}.
We use our B/4 + Ti/16 model variant, and vary the spatial resolution of the tubelets.
Our results indicate that the accuracy is primarily impacted by the spatial resolution of the large encoder. We also note that processing more tokens, and thus using more computation, typically results in higher accuracies.


\begin{table}[t]
    \centering
    \caption{
    Effect of spatial resolution of tubelets.
    All experiments are conducted on Kinetics 400 using the model variant B/4+Ti/16.
    Accuracies are for $4 \times 3$ crops.}
    \footnotesize{
    \begin{tabular}{cccc}
    \toprule
    \multicolumn{2}{c}{Tubelet spatial size} & GFLOPs & Top-1  \\ \midrule
    B & Ti & & \\
    $24\times 24$ & $16 \times 16$    & 68 & 78.1 \\
    $16\times 16$ & $24 \times 24$    & 165 & 80.5 \\
    $16\times 16$ & $16 \times 16$    & 168 & 80.5 \\
    $16\times 16$ & $12 \times 12$    & 169 & 80.6 \\
    $12\times 12$ & $16 \times 16$    & 295 & 81.0 \\
    \bottomrule
    \label{tab:spatial_ablations}
    \end{tabular}
    \vspace{-1.5\baselineskip}
    }
\end{table}

\subsection{Hyperparameters for each datasets} \label{sec: hparams}
\input{hparams}
Table~\ref{tab:training_hyperparameters} details the hyperparamters used in all of our experiments.
We use synchronous SGD with momentum, a cosine learning rate schedule with linear warmup, and a batch size of 64 for all experiments on the Kinetic datasets.
We found that larger batch size and additional regularization are helpful when training on the smaller Epic Kitchens and Something-Something v2 datasets, as also noted by~\cite{arnab2021vivit}.

\subsection{Model configurations} \label{sec: model_configs}
Table \ref{tab:model_configs} summarizes our model configurations of each view for our multiview transformer encoder. For the backbone of each view, we consider five ViT variants, ``Tiny'', ``Small'', ``Base'', ``Large'', and ``Huge''. Their settings strictly follow the ones defined in BERT~\cite{devlin_naacl_2019} and ViT~\cite{dosovitskiy2020image,steiner2021train}. For the global encoder, all model variants of MTV use the same global encoder which follows the “Base” architecture, except that the number of heads is set to 8 instead of 12.
The reason is that the hidden dimension of the tokens should be divisible by the number of heads for multi-head attention, and the number of hidden dimensions across all backbone sizes is divisible by 8 (as shown in Tab.~\ref{tab:model_configs}).
All model variants of MTV (unfactorized) use a one-layer MLP with the same hidden dimension as the “Base” architecture.

\begin{table*}[t]
	\centering
	\caption{Model configurations for each view of MTV.}
		\begin{tabularx}{\linewidth}{l Y Y Y Y Y}
			\toprule
			 Model name & Hidden size & MLP dimension & Number of attention heads  & Number of encoder layers & Tubelet spatial size \\ \midrule
		    Tiny  &     192  &   768   &  3  & 12  &  16   \\			%
			Small & 	384	 &   1536  &  6  & 12  &  16   \\  %
			Base  & 	768	 &   3072  &  12  & 12  &  16   \\  %
			Large & 	1024 &   4096  &  16  & 24  &  16   \\  %
			Huge  &  	1280 &   5120  &  16  & 32  &  14   \\  %
			\bottomrule
		\end{tabularx}
		\label{tab:model_configs}
\end{table*}

%% file: hparams.tex
\begin{table*}[t]
\caption{Training hyperparamters for experiments in the main paper.  ``--'' indicates that the regularisation method was not used at all. 
Values which are constant across all columns are listed once.
Datasets are denoted as follows: K400: Kinetics 400. K600: Kinetics 600. K700: Kinetics 700. MiT: Moments in Time. EK: Epic Kitchens. SSv2: Something-Something v2.}
\begin{tabularx}{\linewidth}{lYYYYYY} %
\toprule
                              & K400 & K600 & K700 & MiT & EK & SSv2 \\ \midrule
\multicolumn{7}{l}{\textit{Optimization}}                                                                                 \\
Optimizer               & \multicolumn{6}{c}{Synchronous SGD}                                                               \\
Momentum					& \multicolumn{6}{c}{0.9} \\
Batch size & 64 & 64 & 64 & 256 & 128 & 512 \\
Learning rate schedule	& \multicolumn{6}{c}{cosine with linear warmup} \\
Linear warmup epochs	& \multicolumn{6}{c}{2.5} \\
Base learning rate			& 0.1 & 0.1 & 0.1 & 0.1 & 0.2 & 0.5 \\
Epochs	& 30 & 30 & 30 & 30 & 80 & 100\\
\midrule
\multicolumn{7}{l}{\textit{Data augmentation}} \\
Random crop probability & \multicolumn{6}{c}{1.0} \\
Random flip probability & 0.5 & 0.5 & 0.5 & 0.5 & 0.5 & -- \\
Scale jitter probability	& \multicolumn{6}{c}{1.0} \\
Maximum scale			  & \multicolumn{6}{c}{1.33} \\
Minimum scale			  & \multicolumn{6}{c}{0.9} \\
Colour jitter probability  & 0.8 & 0.8 & 0.8 & 0.8 & -- & -- \\
Rand augment number of layers~\cite{cubuk_arxiv_2019}		 & -- & -- & -- & -- & 3 & 1 \\
Rand augment magnitude~\cite{cubuk_arxiv_2019} & -- & -- & -- & -- & 10 & 15  \\
\midrule
\multicolumn{7}{l}{\textit{Other regularisation}} \\
Stochastic droplayer rate~\cite{huang_stochasticdepth_eccv_2016} & 0.1 & 0.1 & 0.1 & 0.1 & 0.1 & 0.3 \\
Label smoothing~\cite{szegedy_cvpr_2016}				& -- & -- & -- & -- & 0.2 & 0.2 \\
Mixup~\cite{zhang_mixup_iclr_2018}				 & -- & -- & -- & -- & 0.1 & 0.3 \\
\bottomrule
\end{tabularx}
\label{tab:training_hyperparameters}
\end{table*}